\documentclass{article} %
\usepackage{iclr2023_conference,times}

\usepackage{amsmath,amsfonts,bm}

\def\eqref#1{equation~\ref{#1}}

\def\1{\bm{1}}

\DeclareMathAlphabet{\mathsfit}{\encodingdefault}{\sfdefault}{m}{sl}
\SetMathAlphabet{\mathsfit}{bold}{\encodingdefault}{\sfdefault}{bx}{n}

\definecolor{Gray}{gray}{0.9}
\definecolor{mygreen}{rgb}{0.0, 0.5, 0.0}
\definecolor{myred}{rgb}{0.8, 0.25, 0.33}
\definecolor{myblue}{rgb}{0.19, 0.55, 0.91}
\definecolor{uclablue}{rgb}{0.15, 0.45, 0.68}
\definecolor{boxgreen}{rgb}{0.02, 0.66, 0.02}
\definecolor{boxred}{rgb}{0.66, 0.1, 0.1}
\definecolor{boxblue}{rgb}{0.01, 0.01, 0.73}

\usepackage{url}
\usepackage{etoc}
\usepackage{amsmath}
\usepackage{booktabs}
\usepackage{algorithm}
\usepackage{algorithmic}
\usepackage{lipsum}
\usepackage{pifont}
\usepackage{wrapfig}
\usepackage{colortbl}
\usepackage{acronym}
\usepackage{multicol}
\usepackage{multirow}
\usepackage{makecell}
\usepackage{enumitem}
\usepackage{tabularx,colortbl,array}
\usepackage[skip=3pt,font=small]{subcaption}
\usepackage[skip=3pt,font=small]{caption}
\usepackage[pagebackref,breaklinks,citecolor=uclablue,colorlinks=true,linkcolor=black]{hyperref}
\usepackage{xspace}
\usepackage{mdframed}

\usepackage{mathtools}
\usepackage{amssymb}
\usepackage{amsthm}

\renewcommand{\paragraph}[1]{\noindent\textbf{#1.}}

\newcommand{\smem}{\ensuremath{\mathcal{S}}}
\newcommand{\stxt}{\ensuremath{s^{\text{txt}}}}
\newcommand{\spos}{\ensuremath{s^{\text{pos}}}}
\newcommand{\srot}{\ensuremath{s^{\text{rot}}}}
\newcommand{\question}{\ensuremath{q}}
\newcommand{\answer}{\ensuremath{a}}
\newcommand{\supp}{\textit{appendix}}
\newcommand{\loss}{\mathcal{L}}

\makeatletter
\DeclareRobustCommand\onedot{\futurelet\@let@token\@onedot}
\def\@onedot{\ifx\@let@token.\else.\null\fi\xspace}
\def\eg{\emph{e.g}\onedot} 

\def\ie{\emph{i.e}\onedot}

\def\etc{\emph{etc}\onedot} 

\def\wrt{w.r.t\onedot}

\makeatother

\acrodef{sqa3d}[SQA3D]{\underline{S}ituated \underline{Q}uestion \underline{A}nswering in \underline{3D} Scenes}
\acrodef{vqa}[VQA]{Visual Question Answering}
\acrodef{fpv}[POV]{point of view}
\acrodef{amt}[AMT]{Amazon MTurk}
\acrodef{llms}[LLMs]{Large Language Models}
\acrodef{bev}[BEV]{bird-eye view}

\newcommand{\agentarrow}{\vcenter{\hbox{\includegraphics[width=0.03\textwidth]{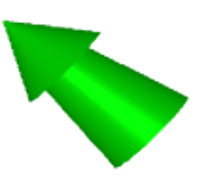}}}}

\usepackage{xcolor}

\definecolor{mygray}{gray}{0.4}
\newcommand{\cmark}{\color{mygray}\ding{51}}%
\newcommand{\xmark}{\color{mygray}\ding{55}}%

\usepackage{soul}

\title{SQA3D:\\ \underline{S}ituated \underline{Q}uestion \underline{A}nswering in \underline{3D} Scenes}

\iclrfinalcopy

\author{Xiaojian Ma$^{2*}$ , Silong Yong$^{1,3}$\thanks{First two authors contributed equally. Correspondence to Zilong Zheng and Siyuan Huang.} , Zilong Zheng$^1$ , Qing Li$^1$ , Yitao Liang$^{1,4}$ \\ \textbf{Song-Chun Zhu$^{1,2,3,4}$ , Siyuan Huang$^{1}$}  \\
$^1$Beijing Institute for General Artificial Intelligence (BIGAI)~~~$^2$UCLA~~~$^3$Tsinghua University\\$^4$Peking University \\
\texttt{\small{xiaojian.ma@ucla.edu, yongzl19@mails.tsinghua.edu.cn}} \\
\texttt{\small{\{zlzheng,liqing,sczhu,syhuang\}@bigai.ai}, yitaol@pku.edu.cn}
}

\begin{document}
\maketitle

\begin{figure}[h]
  \vskip -0.2in
  \begin{center}
    \includegraphics[width=\textwidth]{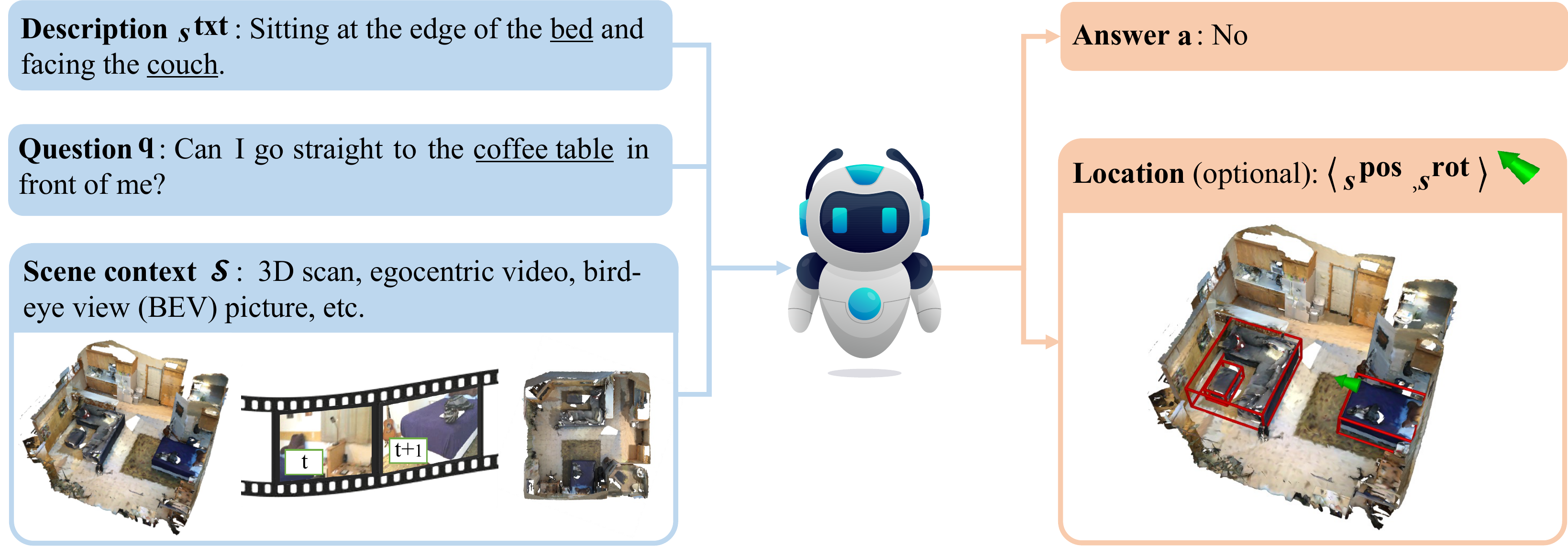}
  \end{center}
  \caption{Task illustration of \underline{S}ituated \underline{Q}uestion \underline{A}nswering in \underline{3D} Scenes~(SQA3D). Given scene context $\smem$ (\eg, 3D scan, egocentric video, bird-eye view picture), SQA3D requires an agent to first comprehend and localize its \textbf{situation} (position, orientation, \etc) in the 3D scene from a textual description $\stxt$, then answer a question $\question$ under that situation. \textbf{Note that understanding the situation and imagining the corresponding egocentric view correctly is necessary to accomplish our task.} %
  We provide more example questions in \autoref{fig:data_example}.
  }
  \label{fig:teaser}
  \vskip -0.12in
\end{figure}

\begin{abstract}

We propose a new task to benchmark scene understanding of embodied agents: \ac{sqa3d}.
Given a scene context (\eg, 3D scan), SQA3D requires the tested agent to first understand its \textbf{situation} (position, orientation, etc.) in the 3D scene as described by text, then reason about its surrounding environment and answer a question under that situation. 
Based upon 650 scenes from ScanNet, we provide a dataset centered around 6.8k unique situations, along with 20.4k descriptions and 33.4k diverse reasoning questions for these situations.
These questions examine a wide spectrum of reasoning capabilities for an intelligent agent, ranging from spatial relation comprehension to commonsense understanding, navigation, and multi-hop reasoning. SQA3D imposes a significant challenge to current multi-modal especially 3D reasoning models. We evaluate various state-of-the-art approaches and find that the best one only achieves an overall score of 47.20\%, while amateur human participants can reach 90.06\%. We believe SQA3D could facilitate future embodied AI research with stronger situation understanding and reasoning capabilities. Code and data are released at \href{https://sqa3d.github.io}{sqa3d.github.io}.

\end{abstract}

\section{Introduction} \label{sec:intro}

In recent years, the endeavor of building intelligent embodied agents has delivered fruitful achievements. Robots now can navigate~\citep{anderson2018vln} and manipulate objects~\citep{liang2019pointnetgpd,savva2019habitat,shridhar2022cliport,ahn2022can} following natural language commands or dialogues. Albeit these promising advances, their actual performances in real-world embodied environments could still fall short of human expectations, especially in generalization to different situations (scenes and locations) and tasks that require substantial, knowledge-intensive reasoning.     
To diagnose the fundamental capability of realistic embodied agents, we investigate the problem of \textbf{embodied scene understanding}, where the agent needs to understand its situation and the surroundings in the environment from a \textit{dynamic} egocentric view, then perceive, reason, and act accordingly, to accomplish complex tasks. 

\textbf{What is at the core of embodied scene understanding?} Drawing inspirations from situated cognition~\citep{greeno1998situativity,anderson2000perspectives}, a seminal theory of embodiment, we anticipate it to be two-fold:
\begin{itemize}[leftmargin=*,noitemsep,topsep=0pt]
    \item \textbf{Situation understanding.} The ability to imagine what the agent will see from arbitrary situations (position, orientations, \etc) in a 3D scene and understand the surroundings anchored to the situation, therefore generalize to novel positions or scenes; 
    \item \textbf{Situated reasoning.} The ability to acquire knowledge about the environment based on the agents' current situation and reason with the knowledge, therefore further facilitates accomplishing complex action planning tasks. 
\end{itemize}

\begin{figure}[t!]
    \centering
    \vskip -0.2in
    \caption{\textbf{Examples from \ac{sqa3d}.} We provide some example questions and the corresponding situations ($\stxt$ and $\agentarrow$) and 3D scenes. The categories listed here do not mean to be exhaustive and a question could fall into multiple categories. The \textcolor{boxgreen}{green boxes} indicate relevant objects in situation description $\stxt$ while \textcolor{boxred}{red boxes} are for the questions $q$.}
    \includegraphics[width=\linewidth]{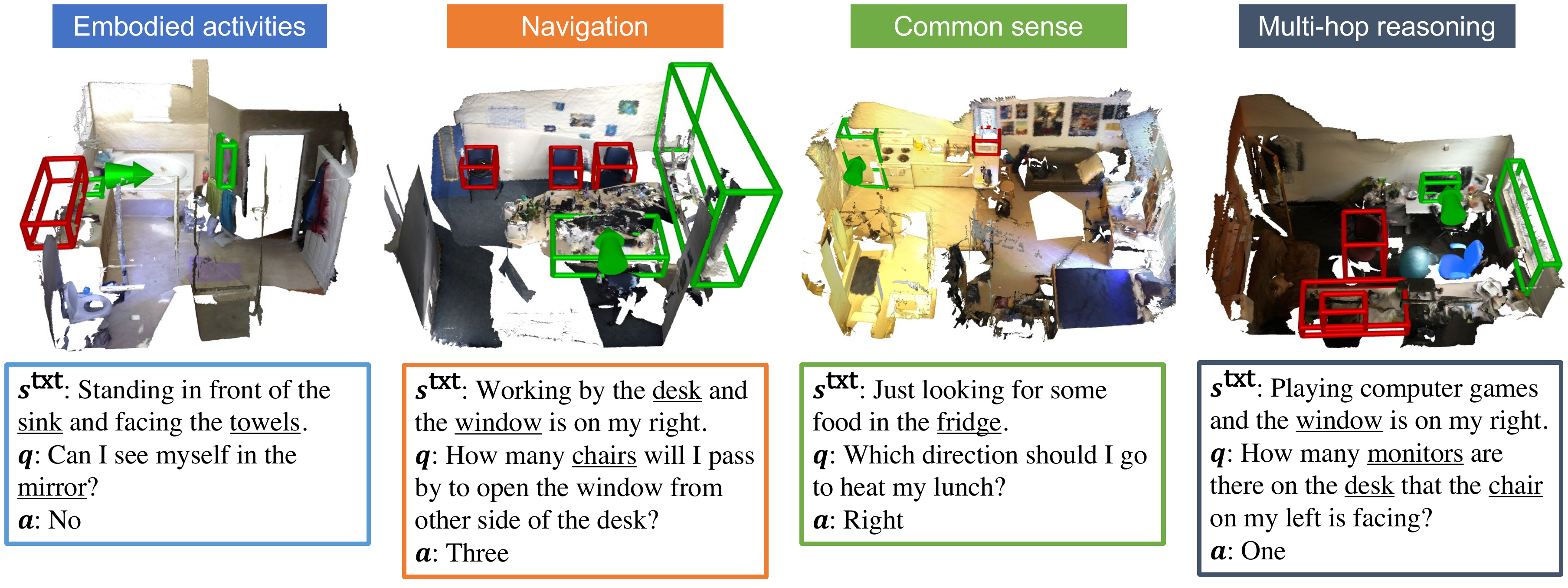}
    \label{fig:data_example}
    \vspace{-0.5in}
\end{figure}

To step towards embodied scene understanding, we introduce \textbf{\ac{sqa3d}}, a new task that reconciles the best of both parties, situation understanding, and situated reasoning, into embodied 3D scene understanding. \autoref{fig:teaser} sketches our task: given a 3D scene context (\eg, 3D scan, ego-centric video, or \ac{bev} picture), the agent in the 3D scene needs to first comprehend and localize its situation (position, orientation, \etc) from a textual description, then answer a question that requires substantial situated reasoning from that perspective. We crowd-sourced the situation descriptions from \ac{amt}, where participants are instructed to select diverse locations and orientations in 3D scenes. To systematically examine the agent's ability in situated reasoning, we collect questions that cover a wide spectrum of knowledge, ranging from spatial relations to navigation, common sense reasoning, and multi-hop reasoning. In total, \ac{sqa3d} comprises 20.4k descriptions of 6.8k unique situations collected from 650 ScanNet scenes and 33.4k questions about these situations. Examples of \ac{sqa3d} can be found \autoref{fig:data_example}.

Our task closely connects to the recent efforts on 3D language grounding~\citep{dai2017scannet,chen2020scanrefer,chen2021scan2cap,hong2021vlgrammar,achlioptas2020referit3d,wang2022humanise,azuma2022scanqa}. However, most of these avenues assume observations of a 3D scene are made from some third-person perspectives rather than an embodied, egocentric view, and they primarily inspect \textit{spatial understanding}, while \ac{sqa3d} examines scene understanding with a wide range of knowledge, and the problems have to be solved using an (imagined) first-person view. Embodied QA~\citep{das2018embodied,wijmans2019eqa} draws very similar motivation as \ac{sqa3d}, but our task adopts a simplified protocol (QA only) while still preserving the function of benchmarking embodied scene understanding, therefore allowing more complex, knowledge-intensive questions and a much larger scale of data collection. Comparisons with relevant tasks and benchmarks are listed in \autoref{tab:comp}.

\textbf{Benchmarking existing baselines}: In our experiments, we examine state-of-the-art multi-modal reasoning models, including ScanQA from~\cite{azuma2022scanqa} that leverages 3D scan data, ClipBERT~\citep{lei2021less} and MCAN~\citep{yu2019deep} that exploits egocentric videos and BEV pictures. However, the results unveil that both models still largely fall behind human performances by a large margin (47.2\% of the best model vs. 90.06\% of amateur human testers). To understand the failure modes, we conduct experiments on settings that could alleviate the challenges brought by situation understanding. The improvement of these models confirms that the current models are indeed struggling with situation understanding, which is pivotal for embodied scene understanding. Finally, we explore whether powerful \ac{llms} like GPT-3~\citep{brown2020language} and Unified QA~\citep{khashabi2020unifiedqa} could tackle our tasks by converting the multi-modal \ac{sqa3d} problems into single-modal surrogates using scene captioning. However, our results read that these models can still be bottlenecked by the lack of spatial understanding and accurate captions.

Our contributions can be summarized as follow: 
\setlength{\leftmargini}{0.85em}
\begin{itemize}[topsep=0pt]
\item We introduce \ac{sqa3d}, a new benchmark for embodied scene understanding, aiming at reconciling the challenging capabilities of situation understanding and situated reasoning and facilitating the development of intelligent embodied agents.
\item We meticulously curate the \ac{sqa3d} to include diverse situations and interesting questions. These questions probe a wide spectrum of knowledge and reasoning abilities of embodied agents, ranging from spatial relation comprehension to navigation, common sense reasoning, and multi-hop reasoning. 
\item We perform extensive analysis on the state-of-the-art multi-modal reasoning models. However, experimental results indicate that these avenues are still struggling on \ac{sqa3d}. 
Our hypothesis suggests the crucial role of proper 3D representations and the demand for better situation understanding in embodied scene understanding.
\end{itemize}

\begin{table}[t!]
\vskip -0.2in
    \caption{\textbf{An overview of the different benchmark datasets covering grounded 3D scene understanding}. In general, we consider semantic grounding, language-driven navigation, and question-answering in photo-realistic 3D scenes. In the first row, \textit{situated} indicates whether the benchmark task is supposed to be completed by a ``situated'' agent with its egocentric perspective. \textit{navigation}, \textit{common sense}, and \textit{multi-hop reasoning} show whether the task requires a certain capability or knowledge level of 3D understanding. $^*$Rather than observing a complete 3D scan of the scene, the learner needs to navigate in a simulator to perceive the 3D scene incrementally.}
    \small{
    \centering
    \resizebox{\linewidth}{!}{%
    \begin{tabular}{c|c|c|c|c|c|c|c|c|c}
    \toprule
    \multirow{2}{*}{dataset} & \multirow{2}{*}{task} & \multirow{2}{*}{situated?} & 3D & text & navi- & common & multi-hop&\multirow{2}{*}{\#scenes} & \multirow{2}{*}{\#tasks} \\
    & &  & type & collection & gation? & sense? & reasoning?& &  \\
    \midrule
    ScanNet~\citep{dai2017scannet} & seg. & \xmark & scan & n/a & \xmark & \xmark & \xmark & 800 rooms & 1.5k  \\
    \midrule
    ScanRefer~\citep{chen2020scanrefer} & det. & \xmark &scan & human & \xmark & \xmark & \xmark & 800 rooms  & 52k \\
    ReferIt3D~\citep{achlioptas2020referit3d} & det. & \xmark & scan & human & \xmark & \xmark & \xmark & 707 rooms & 41k\\
    \midrule
    ScanQA~\citep{azuma2022scanqa}  & q.a. &\xmark & scan & template & \xmark & \xmark & \xmark & 800 rooms & 41k \\
    3D-QA~\citep{ye20213dqa}  & q.a. &\xmark & scan & human & \xmark & \xmark & \xmark & 806 rooms & 5.8k \\ 
    CLEVR3D~\citep{yan2021clevr3d}  & q.a. &\xmark & scan & template & \xmark & \xmark  & \cmark & 478 rooms &60k\\
    \midrule
    MP3D-R2R~\citep{anderson2018vln} & nav. & \cmark & $^*$nav. & human & \cmark & \xmark & \xmark & 190 floors &  22k\\
    MP3D-EQA~\citep{wijmans2019eqa} & q.a. &\cmark & $^*$nav. & template & \cmark & \xmark & \xmark & 146 floors & 1.1k \\
    \midrule
    \ac{sqa3d} (Ours) & q.a. & \cmark & scan & human & \cmark & \cmark & \cmark & 650 rooms & 33.4k \\
    \bottomrule
    \end{tabular}
    }
    }
    \label{tab:comp}
    \vskip -0.2in
\end{table}

\section{The \ac{sqa3d} Dataset}
 \vskip-0.1in

A problem instance in \ac{sqa3d} can be formulated as a triplet $\langle \smem, s, \question \rangle$, where $\smem$ denotes the scene context, \eg, 3D scan, egocentric video, \acf{bev} picture, \etc; $s = \langle \stxt, \spos, \srot \rangle$ denotes a situation, where the textual situation description $\stxt$ (\eg, ``\textit{Sitting at the edge of the bed and facing the couch}'' in \autoref{fig:teaser}) depicts the position $\spos$ and orientation $\srot$ of an agent in the scene; Note that the agent is assumed to be first rotated according to \srot at the origin of the scene coordinate and then translated to \spos; $\question$ denotes a question. The task is to retrieve the correct answer from the answer set $\answer = \{\answer_1, \dots, \answer_N\}$, while optionally predicting the ground truth location $\langle\spos, \srot\rangle$ from the text. The additional prediction of location could help alleviate the challenges brought by situation understanding.
The following subsections will detail how to collect and curate the data and then build the benchmark.

\subsection{Data Formation}\label{sec:data_formation}
 \vskip-0.1in
The 3D indoor scenes are selected from the ScanNet~\citep{dai2017scannet} dataset. We notice that some scenes could be too crowded/sparse, or overall tiny, making situations and questions collection infeasible. Therefore, we first manually categorize these scenes based on the richness of objects/layouts and the space volume. We end up retaining 650 scenes after dropping those that failed to meet the requirement. We then develop an interactive web-based user interface~(UI) to collect the data. Details of UI design can be found in \supp. All the participants are recruited on \ac{amt}.

\begin{figure}[t!]
    \centering
    \vskip -0.4in
    \includegraphics[width=\linewidth]{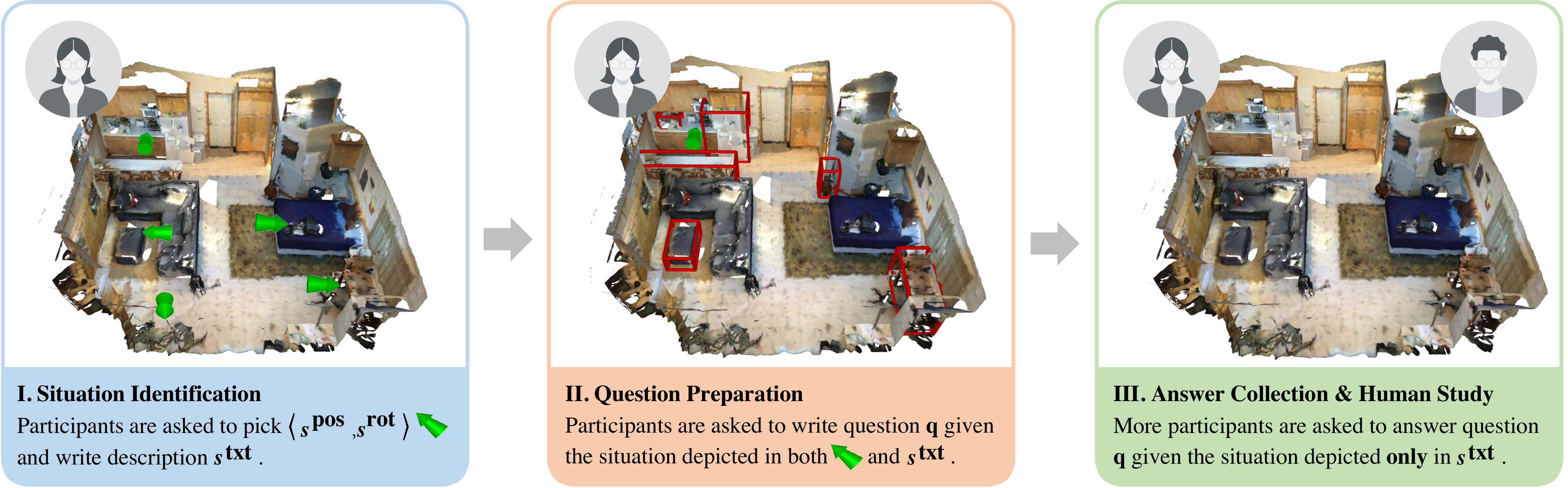}
    \caption{\textbf{Data collection pipeline of \ac{sqa3d}.} Since our dataset comprises multiple types of annotations (situations and their descriptions, questions, answers, \etc), we found it more manageable to break down a single annotation task into three sub-tasks: i) Situation Identification; ii) Question Preparation; iii) Answer Collection \& Human Study, where the participants recruited on \ac{amt} only need to focus on a relatively simple sub-task at a time.}
    \label{fig:qa_collection}
    \vskip -0.2in
\end{figure}

Compared to counterparts, the annotation load of a single \ac{sqa3d} problem instance could be significantly heavier as participants need to explore the scene, pick a situation, make descriptions, and ask a few questions. All these steps also require dense interaction with the 3D scene. To ensure good quality, we introduce a \textbf{multi-stage collection} pipeline, which breaks down the load into more manageable sub-tasks. \autoref{fig:qa_collection} delineates this process:

\textbf{I. Situation Identification.} We ask the workers to pick 5 situations by changing the location $\langle\spos, \srot\rangle$ of a virtual avatar $\agentarrow$ in a ScanNet scene $\smem$. The workers are then instructed to write descriptions $\stxt$ that can \textbf{uniquely} depict these situations in the scene. We also use examples and bonuses to encourage \textbf{more natural sentences} and the \textbf{use of human activities} (\eg, ``\textit{I'm waiting for my lunch to be heated in front of the microwave}"). All the collected situations are later manually curated to ensure diversity and the least ambiguity. If necessary, we would augment the data with more situations to cover different areas of the scene.

\textbf{II. Question Preparation.} We collect a set of questions \wrt each pair of the 3D scene $\smem$, and the situation description $\stxt$ (the virtual avatar $\agentarrow$ is also rendered at $\langle\spos, \srot\rangle$). To help prepare questions that require \textbf{substantial situated reasoning}, we tutor the workers before granting them access to our tasks. They are instructed to follow the rules and learn from good examples. We also remove \& penalize the responses that do not depend on the current situation, \eg. ``\textit{How many chairs are there in the room?}''.

\textbf{III. Answer Collection \& Human Study.} In addition to the answers collected alongside the questions, we send out the questions to more workers and record their responses. These workers are provided with the same interface as in stage \textbf{II} except showing $\agentarrow$ in the scene to ensure consistency between question and answer collection. There is also \textbf{mandatory scene familiarization} in all three steps before the main job starts and we find it extremely helpful especially for more crowded scenes. More details can be found in \supp.

\subsection{Curation, Data Statistics, and Metrics}
 \vskip-0.1in

\paragraph{Curation}
Our multi-stage collection ends up with around 21k descriptions of 6.8k unique situations and 35k questions. Although the aforementioned prompt did yield many high-quality annotations, some of them are still subject to curation. We first apply a basic grammar check to clean up the language glitches. Then we follow the practices in VQAv2~\citep{goyal2017making} and OK-VQA~\citep{marino2019ok} to further eliminate low-effort descriptions and questions. Specifically, we eliminate \& rewrite template-alike descriptions (\eg, repeating the same sentence patterns) and questions that are too simple or do not require looking at the scene. We also notice the similar answer bias reported in~\citet{marino2019ok} where some types of questions might bias toward certain answers. Therefore, we remove questions to ensure a more uniform answer distribution. A comparison of answer distribution before and after the balancing can be found in \supp. As a result, our final dataset comprises 20.4k descriptions and 33.4k diverse and challenging questions. \autoref{fig:data_example} demonstrates some example questions in \ac{sqa3d}.

\paragraph{Statistics}
Compared to most counterparts with template-based text generation, \ac{sqa3d} is crowd-sourced on \ac{amt} and therefore enjoys more naturalness and better diversity. To the best of our knowledge, \ac{sqa3d} is the \textbf{largest} dataset of grounded 3D scene understanding with the human-annotated question-answering pairs (a comparison to the counterparts can be found in~\autoref{tab:comp}). \autoref{tab:statistics}, Figure \ref{fig:word_cloud}, and Figure \ref{fig:question_prefix} illustrate the basic statistics of our dataset, including the word cloud of situation descriptions and question distribution based on their prefixes. It can be seen that descriptions overall meet our expectations as human activities like ``sitting'' and ``facing'' are among the most common words. Our questions are also more diverse and balanced than our counterparts, where those starting with ``What'' make up more than half of the questions and result in biased questions~\citep{azuma2022scanqa}. More statistics like distributions over answers and length of the text can be found in \supp.

\begin{table}[t!]
\vskip -0.4in
\begin{minipage}{0.45\linewidth}
\centering
\includegraphics[width=\linewidth]{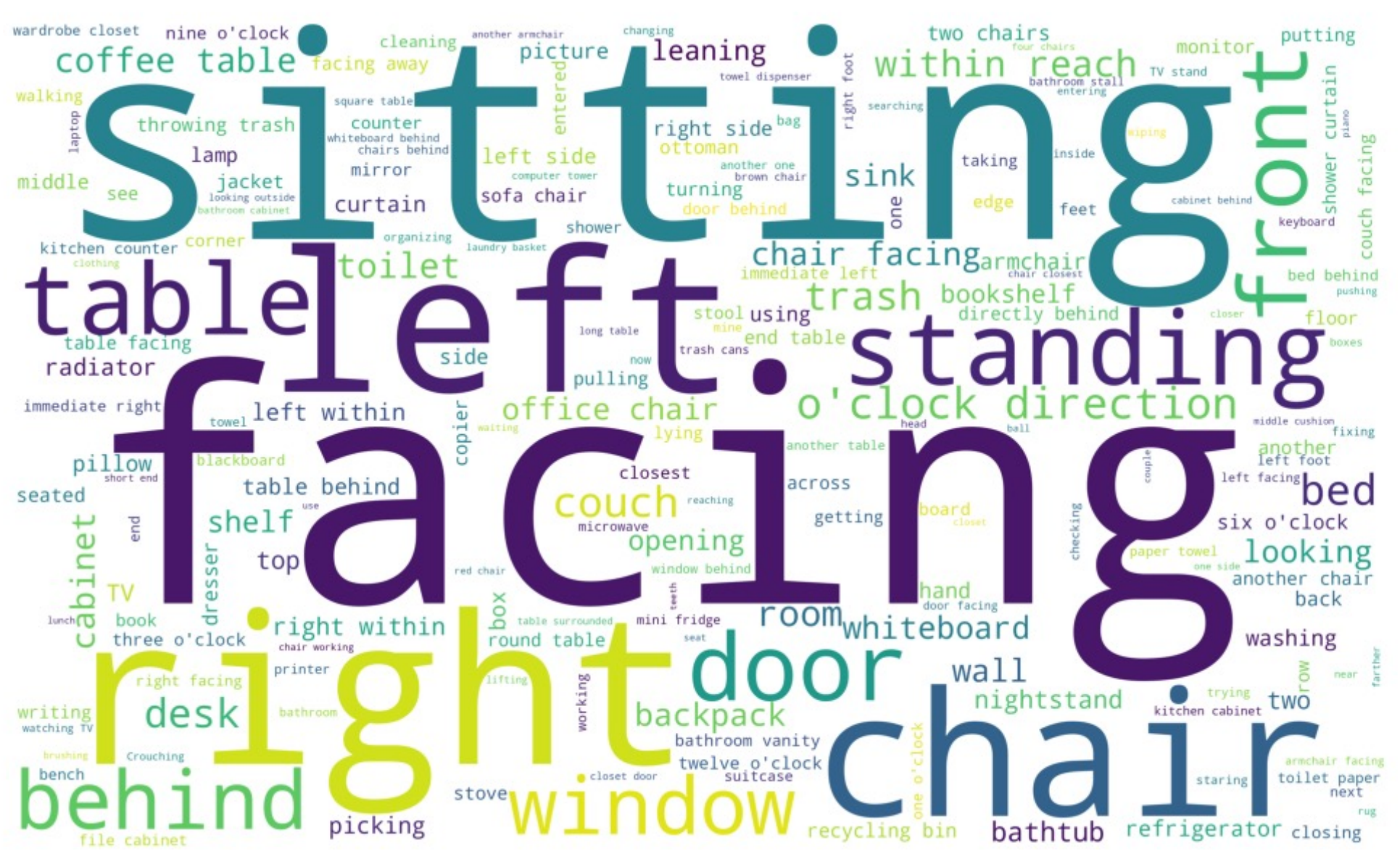}
\captionof{figure}{\label{fig:word_cloud} Word cloud of $\stxt$ in \ac{sqa3d}.}
\vfill
\resizebox{\linewidth}{!}{%
\setlength\tabcolsep{6pt}
\begin{tabular}{lc}
\toprule
     Statistic & Value \\
\midrule
Total $\stxt$ (train/val/test) & 16,229/1,997/2,143\\ 
Total $q$ (train/val/test) & 26,623/3,261/3,519\\
Unique $q$ (train/val/test) & 20,183/2,872/3,036\\
\midrule
Total scenes (train/val/test) & 518/65/67 \\
Total objects (train/val/test) & 11,723/1,550/1,652\\
\midrule
Average $\stxt$ length & 17.49 \\
Average $q$ length &  10.49 \\
\bottomrule
\end{tabular}
}
\captionof{table}{\label{tab:statistics} \ac{sqa3d} dataset statistics.}
\end{minipage}
\hfill
\begin{minipage}{0.55\linewidth}
\centering
    \includegraphics[width=\linewidth]{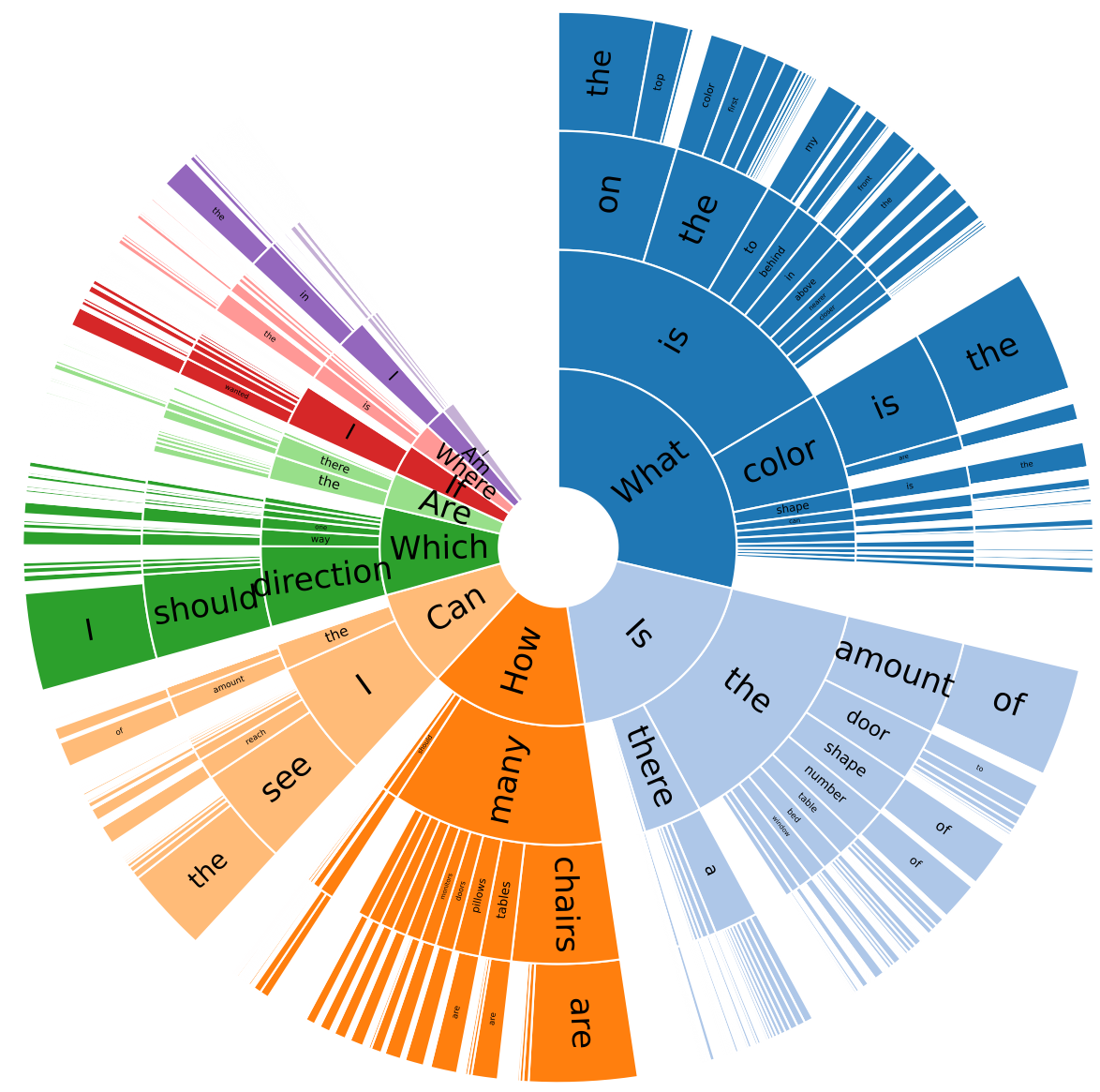}
    \captionof{figure}{\label{fig:question_prefix}Question distribution in \ac{sqa3d}}
\end{minipage}
\vskip -0.2in
\end{table}

\paragraph{Dataset splits and evaluation metric}
We follow the practice of ScanNet and split \ac{sqa3d} into \textit{train}, \textit{val}, and \textit{test} sets. Since we cannot access the semantic annotations in ScanNet \textit{test} set, we instead divide the ScanNet validation scenes into two subsets and use them as our \textit{val} and \textit{test} sets, respectively. The statistics of these splits can be found in \autoref{tab:statistics}. Following the protocol in VQAv2~\citep{goyal2017making}, we provide a set of 706 ``top-K'' answer candidates by excluding answers that only appear very few times. Subsequently, we adopt the ``exact match" as our evaluation metric, \ie, the accuracy of answer classification in the \textit{test} set. No further metric is included as we find it sufficient enough to measure the differences among baseline models with ``exact match".

\section{Models for \ac{sqa3d}}\label{sec:model}
\begin{figure}[t!]
    \centering
    \vskip -0.4in
    \includegraphics[width=\linewidth]{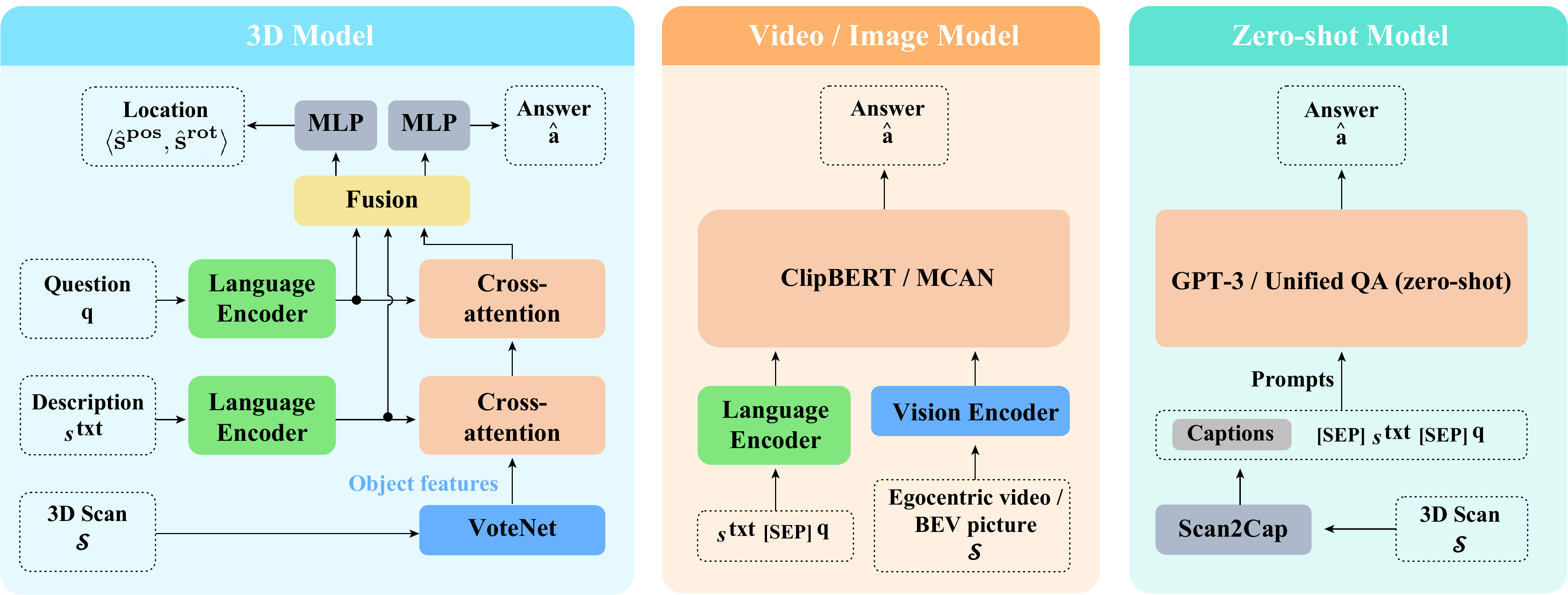}
    \caption{\textbf{Potential models for \ac{sqa3d}.} We split the considered models into three groups: 3D model, video / image model, and zero-shot model. The 3D model is modified from the ScanQA model~\citep{azuma2022scanqa} and maps 3D scan input to the answer. While the video / image models are effectively borrowed from canonical video QA and VQA tasks but we augment them with the additional situation input. The zero-shot model explores the potential of large pre-trained \ac{llms} on our tasks. But they have to work with an additional 3D caption model that converts the 3D scene into text.}
    \label{fig:models}
    \vskip -0.2in
\end{figure}

 \vskip-0.1in
Generally speaking, \ac{sqa3d} can be characterized as a multi-modal reasoning problem. Inspired by the recent advances in transformer-based~\citep{vaswani2017attention} vision-language models~\citep{lu2019vilbert,li2020oscar,alayrac2022flamingo}, we investigate how could these methods approach our task. Specifically, we study a recent transformer-based question-answering system: ScanQA~\citep{azuma2022scanqa}, which maps 3D scans and questions into answers. We make a few adaptations to ensure its compatibility with the protocol in \ac{sqa3d}. To further improve this model, we consider including some auxiliary tasks during training~\citep{ma2022relvit}. For other types of 3D scene context, \eg. egocentric video clips and BEV pictures, we employ the corresponding state-of-the-art models. Finally, we explore the potential of recently-introduced \ac{llms} like GPT-3~\citep{brown2020language} and Unified QA~\citep{khashabi2020unifiedqa} on solving \ac{sqa3d} in a zero-shot fashion. An overview of these models can be found in \autoref{fig:models}.

\paragraph{3D model} We use the term \textit{3D model} to refer a modified version of the ScanQA model~\citep{azuma2022scanqa}, depicted in the blue box of \autoref{fig:models}. It includes a VoteNet~\citep{qi2019deep}-based 3D perception module that extracts object-centric features, LSTM-based language encoders for processing both questions $q$ and situation description $\stxt$, and some cross-attention transformer blocks~\citep{vaswani2017attention}. The object-centric feature tokens attend to the language tokens of $\stxt$ and $q$ successively. Finally, these features will be fused and mapped to predict the answer. Optionally, we can add one head to predict the location $\langle\spos, \srot\rangle$ of the agent. Since the VoteNet module is trained from scratch, we also employ an object detection objective (not shown in the figure).

\paragraph{Auxiliary task} %
As we mentioned before, situation understanding plays a crucial role in accomplishing \ac{sqa3d} tasks. To encourage a better understanding of the specified situation, we introduce two auxiliary tasks: the model is required to make predictions about the $\spos$ and $\srot$ of the situation. We use mean-square-error (MSE) loss for these tasks. The overall loss for our problem therefore becomes $\loss = \loss_{\text{ans}} + \alpha\loss_{\text{pos}} + \beta\loss_{\text{rot}}$, where $\loss_{\text{ans}}$, $\loss_{\text{pos}}$, and $\loss_{\text{rot}}$ depicts the losses of the main and auxiliary tasks, $\alpha$ and $\beta$ are balancing weights.

\paragraph{Video and Image-based model} The orange box in the middle of \autoref{fig:models} demonstrates the models for video and image-based input. \ac{sqa3d} largely resembles a video question answering or visual question answering problem when choosing to represent the 3D scene context $\smem$ as egocentric video clips or BEV pictures. However, \ac{sqa3d} also requires the model to take both question $q$ and the newly added situation description $\stxt$ as input. We, therefore, follow the practice in the task of context-based QA~\citep{rajpurkar2018know} and prepend $\stxt$ to the question as a \textit{context}. For the model, we use the state-of-the-art video QA system ClipBERT~\citep{lei2021less} and VQA system MCAN~\citep{yu2019deep}. We adopt most of their default hyper-parameters and the details can be found in \supp.

\paragraph{Zero-shot model} We explore to which extent the powerful \ac{llms} like GPT-3~\citep{brown2020language} and Unified QA~\citep{khashabi2020unifiedqa} could tackle our tasks. Following prior practices that apply GPT-3 to VQA~\citep{changpinyo2022all,gao2022transform}, we propose to convert the 3D scene into text using an emerging technique called 3D captioning~\citep{chen2021scan2cap}. We provide the caption, $\stxt$, and $q$ as part of the prompt and ask these models to complete the answer. For GPT-3, we further found providing few-shot examples in the prompt helpful with much better results. Minor post-processing is also needed to ensure answer quality. We provide more details on prompt engineering in the \supp.

\section{Experiments}
 \vskip-0.1in

\subsection{Setup}\label{sec:exp_setup}
 \vskip-0.1in

We benchmark the models introduced in \autoref{sec:model} to evaluate their performances on \ac{sqa3d}. As mentioned before, we examine three types of scene context $\smem$: 3D scan (point cloud), egocentric video, and BEV picture. Both the 3D scan and egocentric video for each scene are provided by ScanNet~\citep{dai2017scannet}. However, we down-sample the video to allow more efficient computation per the requirement of the ClipBERT model~\citep{lei2021less}. The BEV pictures are rendered by placing a top-down camera on top of the scan of each 3D scene. We also conduct additional experiments that investigate factors that could contribute to the results, \eg, situation and auxiliary tasks. In our early experiments, we found that the 3D model overall performs better than the video or image-based models. Therefore we only conduct these additional experiments with the variants of our 3D model due to the limit of computational resources. We use the official implementation of ScanQA, ClipBERT, and MCAN and include our modifications for SQA3D. For the zero-shot models, we extract 3D scene captions from two sources: ScanRefer~\citep{chen2020scanrefer} and ReferIt3D~\citep{achlioptas2020referit3d}. Considering the limit on the length of the input prompt, these 3D captions are also down-sampled. The Unified QA model weights are obtained from its Huggingface official repo. All the models are tuned using the validation set and we only report results on the test set. More details on model implementation can be found in \supp.

\begin{table}[t!]
\vskip -0.2in
\setlength\tabcolsep{3pt}
    \centering
    \small
\begin{tabular}{lccccccccc}
\toprule
                    & \multicolumn{1}{c}{\multirow{2}{*}{$\smem$}} & \multicolumn{1}{c}{\multirow{2}{*}{Format}} & \multicolumn{6}{c}{test set} & \multicolumn{1}{c}{\multirow{2}{*}{Avg.}} \\ \cline{4-9}
                    & \multicolumn{1}{c}{}                         & \multicolumn{1}{c}{}                        & What  & Is  & How  & Can & Which & Others  & \multicolumn{1}{c}{}                      \\ \midrule
Blind test          & -           & SQ$\rightarrow$A    &   26.75    &  63.34   & 43.44     & \textbf{69.53}  & 37.89 & 43.41  & 43.65        \\ \midrule
ScanQA (w/o $\stxt$)& 3D scan     & VQ$\rightarrow$A    & 28.58      & 65.03    & \textbf{47.31}     & 66.27   & 43.87 &  42.88 &   45.27      \\
ScanQA       & 3D scan            & VSQ$\rightarrow$A   &  31.64     &  63.80   &  46.02    & \textbf{69.53}  & 43.87 & 45.34  &  46.58   \\
ScanQA + aux. task   & 3D scan    & VSQ$\rightarrow$AL  &  33.48     &  \textbf{66.10}   &  42.37    & \textbf{69.53}   & 43.02 & \textbf{46.40}  &  \textbf{47.20}   \\ \midrule
MCAN                & BEV         & VSQ$\rightarrow$A   & 28.86      & 59.66    & 44.09     & 68.34  & 40.74 & 40.46  & 43.42       \\
ClipBERT            & Ego. video  & VSQ$\rightarrow$A   & 30.24      &  60.12   &  38.71    & 63.31  & 42.45 & 42.71  & 43.31     \\ \midrule
$\text{Unified QA}_{\text{Large}}$ & ScanRefer   & VSQ$\rightarrow$A  & 33.01 &  50.43  & 31.91  & 56.51  & \textbf{45.17} & 41.11  & 41.00   \\
$\text{Unified QA}_{\text{Large}}$ & ReferIt3D   & VSQ$\rightarrow$A  & 27.58 &  47.99  & 34.05  & 59.47  & 40.91 & 39.77  & 38.71   \\
GPT-3   & ScanRefer      & VSQ$\rightarrow$A   &  \textbf{39.67}   & 45.99 &  40.47  & 45.56  & 36.08  & 38.42  & 41.00  \\
GPT-3   & ReferIt3D      & VSQ$\rightarrow$A   &  28.90   & 46.42 &  28.05  & 40.24  & 30.11  & 36.07  & 34.57  \\ \midrule
Human (amateur)     & 3D scan        & VSQ$\rightarrow$A   &  88.53   & 93.84 &  88.44  & 95.27  & 87.22  & 88.57  & 90.06 \\ \bottomrule
\end{tabular}
    \caption{\textbf{Quantitative results on the \ac{sqa3d} benchmark}. Results are presented in accuracy (\%) on different types of questions. In the ``Format'' column: V = 3D visual input $\smem$; S = situation description $\stxt$; Q = question $q$; A = answer $a$; L = location $\langle\spos, \srot\rangle$. In ScanQA, \textit{aux. task} indicates the use of both $\loss_{\text{pos}}$ and $\loss_{\text{rot}}$ as additional losses. We use the \textit{Large} variant as Unified QA~\citep{khashabi2020unifiedqa} as it works better.}
    \label{tab:main_result}
    \vskip -0.3in
\end{table}

\subsection{Quantitative Results}\label{sec:quan_result}
 \vskip-0.1in

We provide the quantitative results of the considered models (detailed in \autoref{sec:model}) on our \ac{sqa3d} benchmark in \autoref{tab:main_result}. The findings are summarized below: 

\paragraph{Question types} In \autoref{tab:main_result}, we demonstrate accuracy on six types of questions based on their prefixes. Most models tend to perform better on the ``Is'' and ``Can'' questions while delivering worse results on ``What'' questions, likely due to a smaller number of answer candidates -- most questions with binary answers start with ``Is'' and ``Can'', offering a better chance for the random guess. Moreover, we observe the hugest gap between the blind test (model w/o 3D scene context input) and our best model on the ``What'' and ``Which'' categories, suggesting the need for more visual information for these two types of questions. This also partially echoes the finding reported in~\cite{lei2018tvqa}.

\paragraph{Situation understanding and reasoning} At the heart of \ac{sqa3d} benchmark is the requirement of situation understanding and reasoning. As we mentioned in~\autoref{sec:data_formation}, the model will be more vulnerable to wrong answer predictions if ignoring the situation that the question depends on (\eg. ``\textit{What is in front of me}'' could have completely different answers under different situations). In~\autoref{tab:main_result}, removing situation description $\stxt$ from the input leads to worse results, while adding the auxiliary situation prediction tasks boosts the overall performance, especially on the challenging ``What'' questions. The only exception is ``How'' questions, where a majority of them are about counting. We hypothesize that most objects in each ScanNet scene only have a relatively small number of instances, and the number could also correlate to the object category. Therefore, guessing/memorization based on the question only could offer better results than models with the situation as input if the situation understanding \& reasoning are still not perfect yet. Additionally, we also provide an inspection of the relation between situation understanding and QA using attention visualization in~\autoref{sec:qual_res}.

\paragraph{Representations of 3D scenes} Indeed, \ac{sqa3d} does not limit the input to be 3D scan only, as we also offer options of egocentric videos and BEV pictures. Compared to models with the 3D scan as input, the tested models with other 3D representations (\ie, MCAN and ClipBERT) deliver much worse results, implying that the 3D scan so far could still be a better representation for the 3D scene when the reasoning models are probed with questions that require a holistic understanding of the scene. On the other hand, MCAN and ClipBERT are general-purpose QA systems, while ScanQA is designed for 3D-language reasoning tasks. The generalist-specialty trade-off could also partially account for the gap. Finally, the poor results of BEV and egocentric videos based models compared to the blind test could also be due to the additional ``vision-bias'' when the visual input is provided~\citep{antol2015vqa}. Note that the vision-bias can be mitigated with better visual representations~\citep{wen2021debiased}, implying that ScanQA, which seems to suffer less from the vision-bias than the counterparts using BEV and egocentric videos, is fueled by better visual representations in terms of combating the dataset bias.

\paragraph{Zero-shot vs. training from scratch} The success of pre-trained \ac{llms} like GPT-3 on myriads of challenging reasoning tasks~\citep{wei2022chain,wei2022emergent} suggests that these models could possibly also understand embodied 3D scenes with language-only input~\citep{landau1993and}. However, \ac{sqa3d} imposes a grand challenge to these models. The powerful Unified QA (\textit{Large} variant) and GPT-3 both fail to deliver reasonable results on our tasks. Further, we hypothesize the bottleneck could also be on the 3D captions, as the results verify the consistent impact on model performances brought by a different source of captions (ScanRefer$\rightarrow$ReferIt3D). However, we still believe these models have great potential. For example, one zero-shot model (GPT-3 + ScanRefer) do pretty well on the challenging ``What'' questions (39.67\%), even better than the best ScanQA variant.

\paragraph{Human vs. machine} Finally, all the machine learning models largely fall behind amateur human participants (47.2\% of ScanQA + aux. task vs. 90.06\%). Notably, we only offer a limited number of examples for the testers before sending them the \ac{sqa3d} problems. Our participants promptly master how to interact with the 3D scene, understand the situation from the textual description, and answer the challenging questions. The human performance also shows no significant bias for different question types.

\begin{figure}[t!]
    \centering
    \vskip -0.2in
    \includegraphics[width=\linewidth]{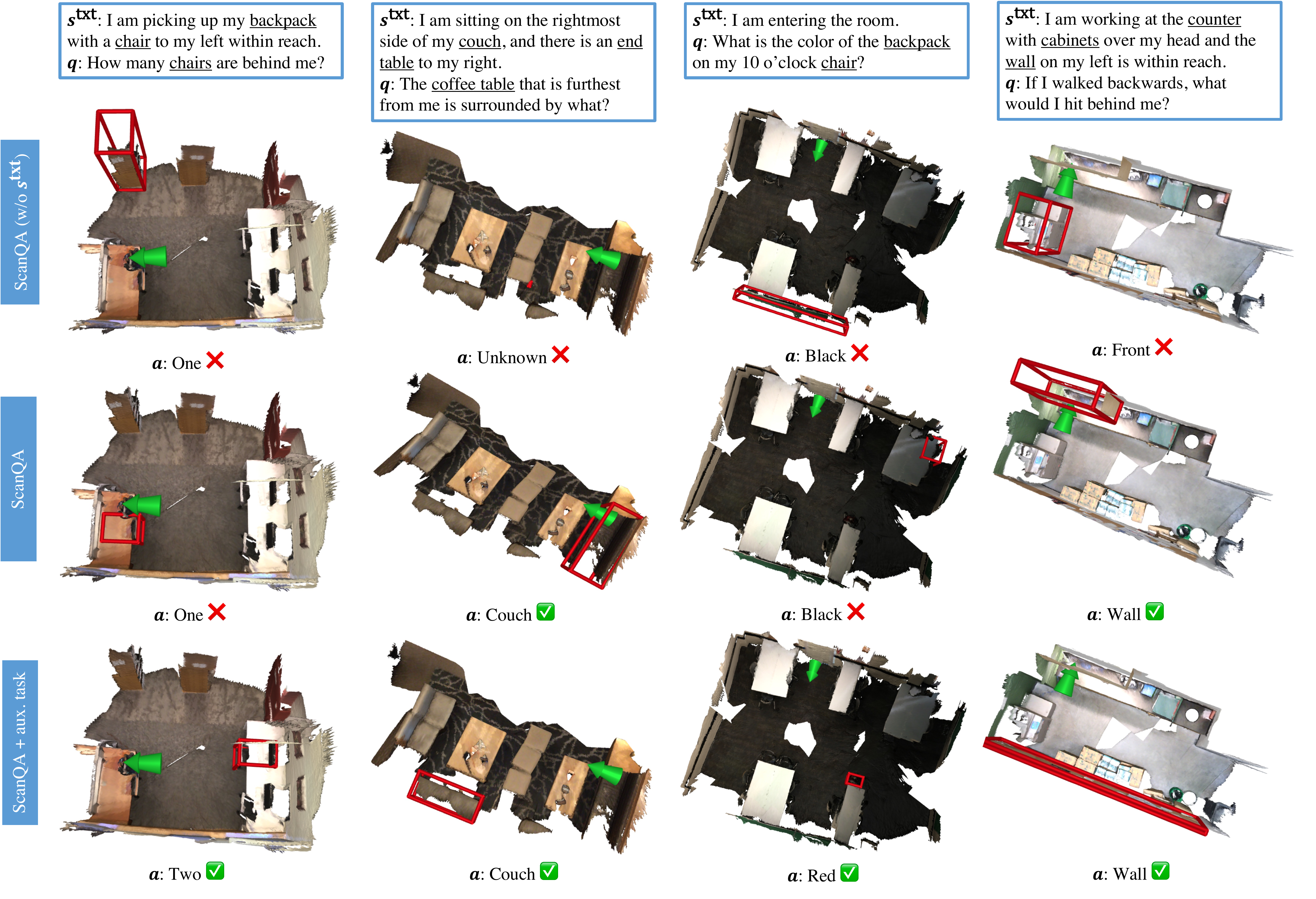}
    \caption{\textbf{Qualitative results.} We show the predicted answer and \textcolor{boxred}{bbox} with highest attention for the variants of ScanQA~\citep{azuma2022scanqa} models. We anticipate the \textcolor{boxred}{bbox} to indicate the object that situation description $\stxt$ or question $q$ refers to. We observe that better situation understanding (via comprehension on $\stxt$ or auxiliary tasks) could result in more reasonable attention over objects, which positively correlates to more robust answer prediction.}
    \label{fig:qual_result}
    \vskip -0.3in
\end{figure}

\subsection{Qualitative Results}\label{sec:qual_res}
 \vskip-0.1in
Finally, we offer some qualitative results of the variants of our 3D model in \autoref{fig:qual_result}. We primarily focus on visualizing both the answer predictions and the transformer attention over the object-centric feature tokens (bounding boxes) generated by the VoteNet~\citep{qi2019deep} backbone. We highlight the most-attended bounding box among all the predictions by the transformer-based model, in the hope of a better understanding of how these models perceive the 3D scene to comprehend the situations and answer the questions. In~\autoref{fig:qual_result}, the correct predictions are always associated with attention over relevant objects in the situation description $\stxt$ and questions. Moreover, in case there are multiple instances of the same object category, it is also crucial to identify the correct instance. For example, only ScanQA + aux. task makes the correct prediction for the first question and also attends to the right chair behind $\agentarrow$, while ScanQA focuses on a wrong instance. These results confirm our findings in~\autoref{sec:quan_result} about the critical role of situation understanding. We also provide some failure modes in \supp.

\subsection{Additional Task for Localization}
\vskip -0.1in
As illustrated in \autoref{fig:teaser}, the agent could optionally predict the current location based off the situation description \stxt and the current 3D scene context \smem. We therefore provide some additional metrics to help evaluate these predictions. Specifically, the agent needs to predict both the current position \spos in 3D coordinate $\langle x, y, z\rangle$ (unit is meter) and orientation in quaternion $\langle x, y, z, w\rangle$. Then these predictions will be evaluated separately using the following metrics:
\begin{itemize}
    \item \textbf{Acc@0.5m}: If the predicted position is within 0.5 meter range to the ground truth position, the prediction will be counted as correct. We then report $\frac{\#\text{\texttt{correctly predicted ground truth}}}{\#\text{\texttt{all ground truth}}}$.
    \item \textbf{Acc@1.0m}: Similar to \textbf{Acc@0.5m} but the range limit is 1.0 meter instead.
    \item \textbf{Acc@15°}: If the prediction orientation is within a 15° range to the ground truth orientation, the prediction will be counted as correct.
    \item \textbf{Acc@30°}: Similar to \textbf{Acc@15°} but the range limit is 30° instead. 
\end{itemize}
Note that, for position prediction, we only consider the predicted $x, y$ and for orientation prediction, only the rotation along $z$-axis counts. We report the result of random prediction below as an reference.
\begin{table}[h!]
    \centering
    \begin{tabular}{ccccc}
    \toprule
         & Acc@0.5m & Acc@1.0m & Acc@15° & Acc@30°  \\
    \midrule
         Random & 14.60 & 34.21 & 22.39 & 42.28 \\
    \bottomrule
    \end{tabular}
    \caption{Random predictions evaluated on the localization task.}
    \label{tab:my_label}
\end{table}

\vskip-0.2in
\section{Related Work}
\vskip-0.1in

\paragraph{Embodied AI} The study of embodied AI~\citep{brooks1990elephants} emerges from the hypothesis of ``\textit{ongoing physical interaction with the environment as the primary source of constraint on the design of intelligent systems}''. To this end, researchers have proposed a myriad of AI tasks to investigate whether intelligence will emerge by acting in virtual or photo-realistic environments. Notable tasks including robotic navigation~\citep{das2018embodied,anderson2018vln,savva2019habitat,chen2019touchdown,wijmans2019dd,qi2020reverie,deitke2022procthor} and vision-based manipulation~\citep{kolve2017ai2,puig2018virtualhome,xie2019vrgym,shridhar2020alfred,shridhar2020alfworld,shridhar2022cliport}. These tasks are made more challenging as instructions or natural-dialogues are further employed as conditions. Sophisticated models have also been developed to tackle these challenges. Earlier endeavors usually comprise multi-modal fusion~\citep{tenenbaum1996separating,perez2018film} and are trained from scratch~\citep{wang2018look,fried2018speaker,wang2019reinforced}, while recent efforts would employ pre-trained models~\citep{pashevich2021episodic,hong2021vln,suglia2021embodied}. However, the agents still suffer from poor generalization to novel and more complex testing tasks~\citep{shridhar2020alfred} compared to results on training tasks. More detailed inspection has still yet to be conducted and it also motivates our \ac{sqa3d} dataset, which investigates one crucial capability that the current embodied agents might need to improve: \textbf{embodied scene understanding}. 

\paragraph{Grounded 3D understanding} Visual grounding has been viewed as a key to connecting human knowledge, which is presumably encoded in our language, to the visual world, so as enable the intelligent agent to better understand and act in the real environment. It is natural to extend this ability to 3D data as it offers more immersive representations of the world. Earlier work has examined word-level grounding with detection and segmentation tasks on 3D data~\citep{gupta2013perceptual,song2014sliding,dai2017scannet,chang2017matterport3d}. Recent research starts to cover sentence-level grounding with complex semantics~\citep{chen2020scanrefer,achlioptas2020referit3d,chen2021scan2cap}. More recently, new benchmarks introduce complex visual reasoning to 3D data~\citep{azuma2022scanqa,ye20213dqa,yan2021clevr3d}. However, these tasks mostly assume a passive, third-person's perspective, while our \ac{sqa3d} requires problem-solving with an egocentric viewpoint. This introduces both challenges and chances for tasks that need a first-person's view, \eg. embodied AI. 

\paragraph{Multi-modal question answering} Building generalist question answering (QA) systems has long been a goal for AI. Along with the progress in multi-modal machine learning, VQA~\citep{antol2015vqa,zhu2016visual7w} pioneers the efforts of facilitating the development of more human-like, multi-modal QA systems. It has been extended with more types of knowledge, \eg. common sense~\citep{zellers2019recognition} and factual knowledge~\citep{marino2019ok}. Recent research has also introduced QA tasks on video~\citep{lei2018tvqa,jia2020lemma,jia2022egotaskqa,grunde2021agqa,wu2021star,datta2022episodic}, and 3D data~\citep{ye20213dqa,azuma2022scanqa,yan2021clevr3d}. We propose the \ac{sqa3d} benchmark also in hope of facilitating multi-modal QA systems with the ability of embodied scene understanding. Notably, models for \ac{sqa3d} could choose their input from a 3D scan, egocentric video, or BEV picture, which makes our dataset compatible with a wide spectrum of existing QA systems.

\vskip-0.1in
\section{Conclusion}
 \vskip-0.1in

We've introduced \ac{sqa3d}, a benchmark that investigates the capability of embodied scene understanding by combining the best of situation understanding and situated reasoning. We carefully curate our dataset to include diverse situations and interesting questions while preserving the relatively large scale (20.4k situation descriptions and ~33.4k questions). Our questions probe a wide spectrum of knowledge and reasoning abilities of embodied agents, notably navigation, common sense, and multi-hop reasoning. We examine many state-of-the-art multi-modal reasoning systems but the gap between the best ML model and human performances so far is still significant. Our findings suggest the crucial role of proper 3D representations and better situation understanding. With \ac{sqa3d}, we hope of fostering research efforts in developing better embodied scene understanding methods and ultimately facilitate the emergence of more intelligent embodied agents. 

\section*{Acknowledgement}
The authors would like to thank Dave Zhenyu Chen for his insightful ScanRefer project and help on data collection, Wenjuan Han for discussions on data collection and model design. This project is supported by National Key R\&D Program of China (2021ZD0150200).

\bibliography{main}
\bibliographystyle{iclr2023_conference}

\newpage
\appendix

\section{Data collection}

\subsection{Data collection Web UI}

\begin{figure}[h]
    \centering
    \includegraphics[width=0.7\linewidth]{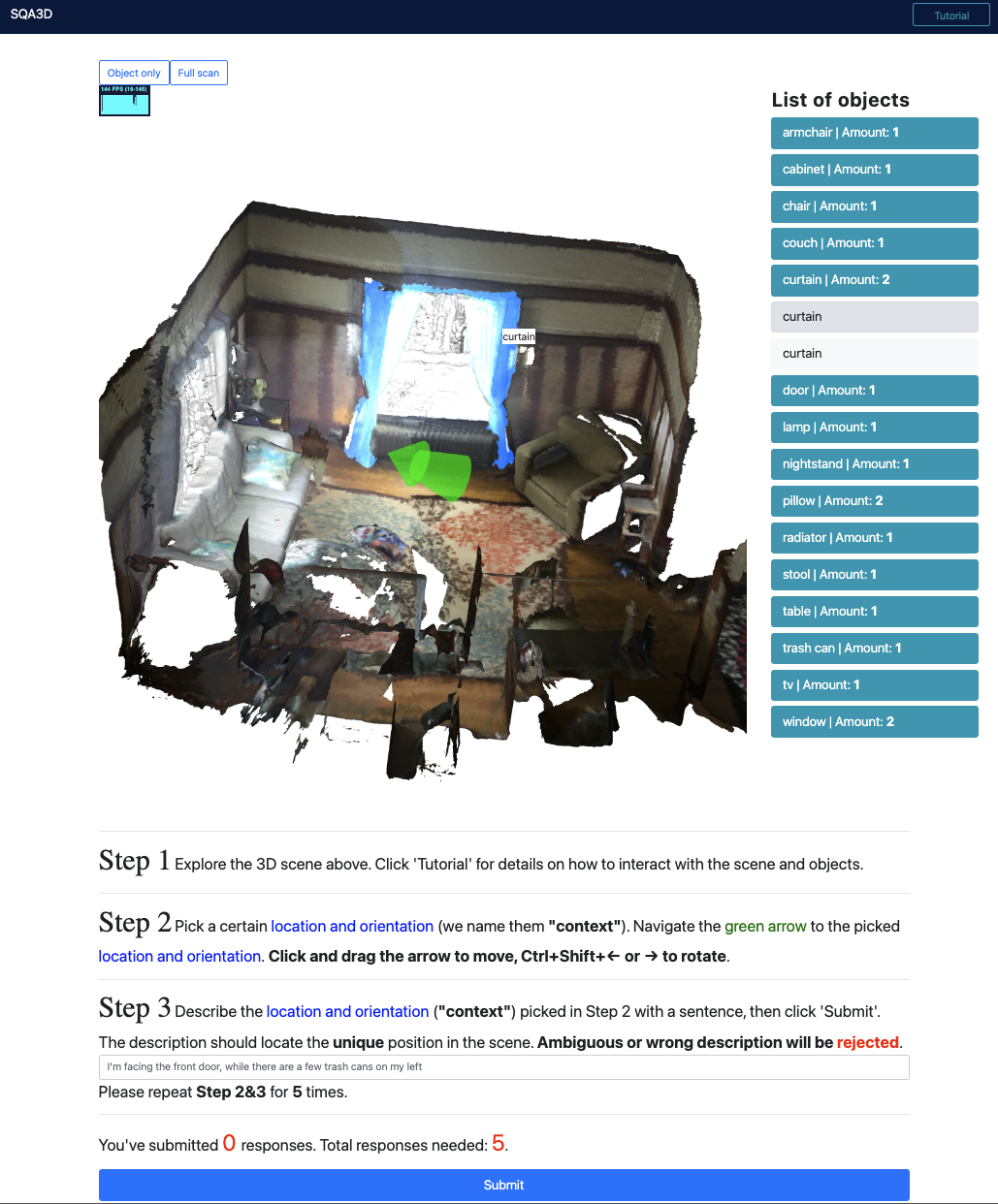}
    \caption{Dataset collection Web UI for Stage I.}
    \label{fig:task1_ui}
\end{figure}

\begin{figure}[h]
    \centering
    \includegraphics[width=0.7\linewidth]{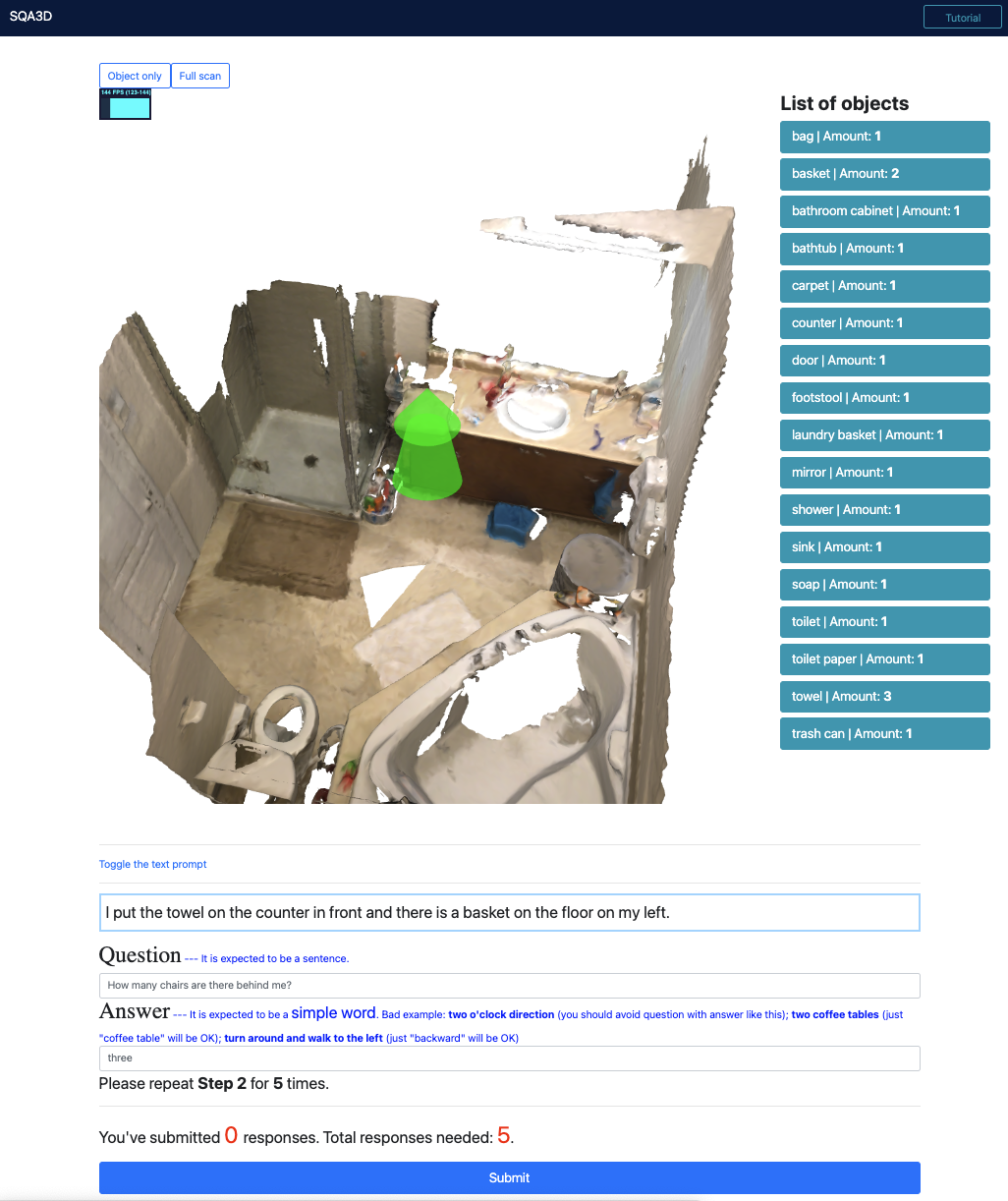}
    \caption{Dataset collection Web UI for Stage II.}
    \label{fig:task2_ui}
\end{figure}

\begin{figure}[h]
    \centering
    \includegraphics[width=0.7\linewidth]{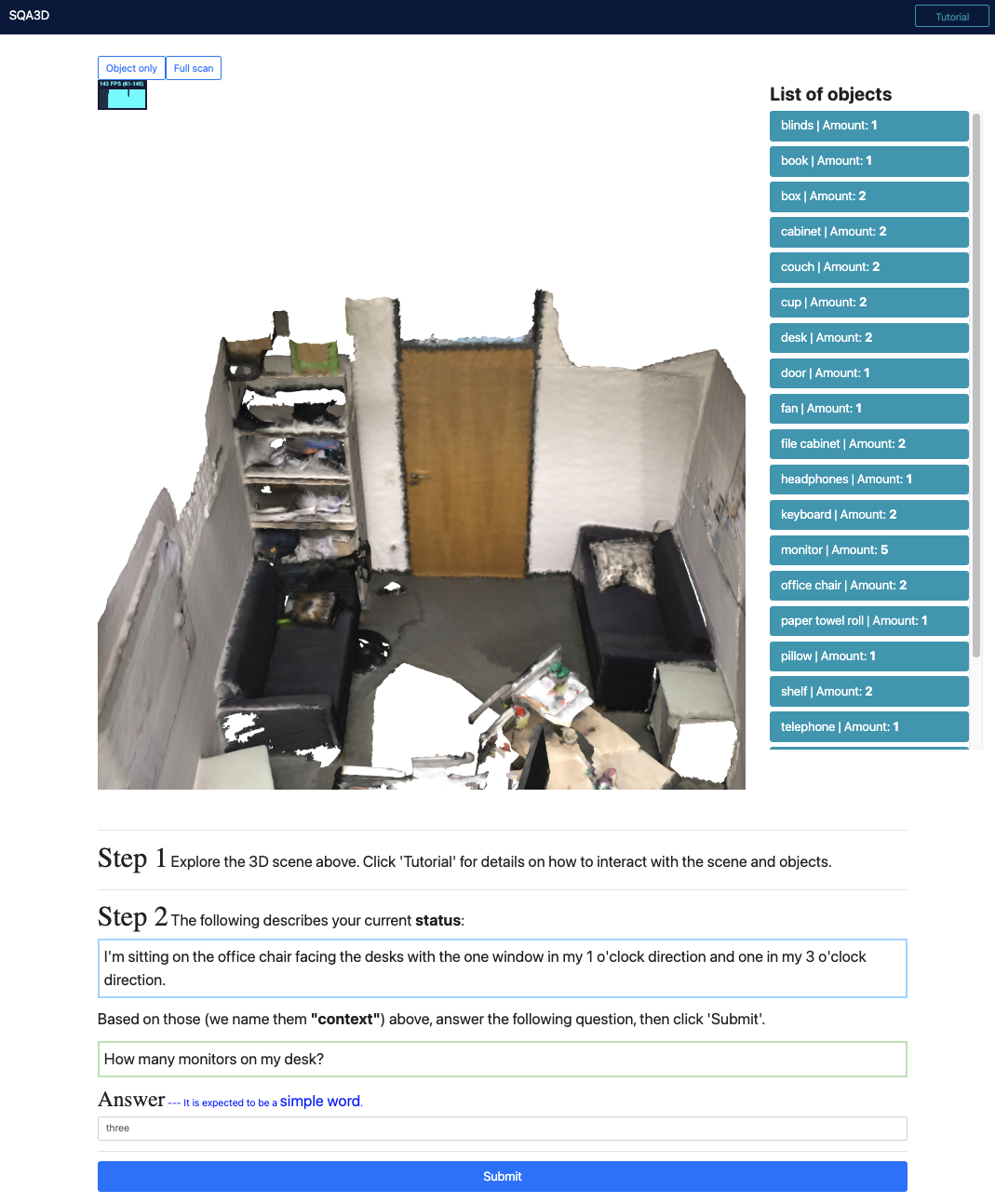}
    \caption{Dataset collection Web UI for Stage III.}
    \label{fig:task3_ui}
\end{figure}

We present the Web UI of our data collection in~\autoref{fig:task1_ui} (Stage I), \autoref{fig:task2_ui} (Stage II) and \autoref{fig:task3_ui} (Stage III) respectively. We developed our UI based on~\cite{chen2020scanrefer}. These UIs share some common components: a 3D scene viewer, where the user can drag, rotate, and zoom in/out the scene; clickable objects/tags, where users might click on either the object mesh directly or the tag on the sidebar to highlight it in the scene; and an instruction set that guide the user through the task. Users may also switch between a full scene or object mesh only to focus on the tasks. The users are also required to submit multiple responses with the same scene.

\begin{figure}[h]
    \centering
    \includegraphics[width=0.8\linewidth]{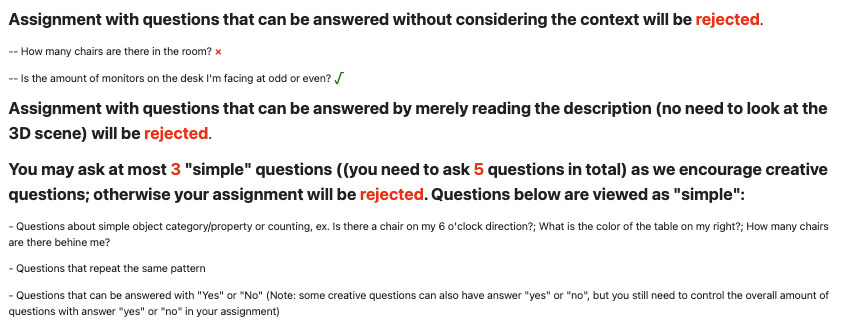}
    \caption{Additional instruction set to the \ac{amt} participants in Stage II.}
    \label{fig:task_ui_reminder}
\end{figure}

Notably, we create detailed tutorials for each stage (not shown in the UI) with examples and animated demonstrations. We found tutorials and instruction sets with clear criteria on \textcolor{myred}{rejection} and \textcolor{mygreen}{bonus}(\eg. \autoref{fig:task_ui_reminder}) helpful with high-quality data. Finally, all the testers need to pass a test before the qualification for our task is granted.

\subsection{Data post-processing}

There are two major data post-processing steps in \ac{sqa3d}: \textbf{cleaning} and \textbf{balancing}. For cleaning, we primarily focus on grammatical correction. We adopt both rule-based cleaning and an ML-based tool called GECToR~\citep{omelianchuk-etal-2020-gector} in our grammatical correction pipeline. We adjust the correction threshold based on human judgment over the corrected data samples.

In the balancing step, our goal is to reduce the question-answer bias in the dataset. Therefore we follow the practice in~\cite{antol2015vqa,marino2019ok} and re-sample the questions based on their prefixes and answer type, in hope of a more balanced answer distribution. We provide answer distribution before and after balancing in~\autoref{sec:more_stat}.

\subsection{More MTurk details}

We provide the detailed MTurk job settings below:

\paragraph{Region} We enable access to our tasks in the following countries/regions:
\begin{table}[h]
\begin{tabular}{|p{35em}|}
\hline
\makecell[tl]{\\US, DE, GB, AU, CA, SG, NZ, NO, SE, FI, DK, IE\\} \\ \hline
\end{tabular}
\end{table}

\paragraph{Approval rate \& Number of approved jobs} The testers are required to have at least a 95\% approval rate and have completed more than 1000 tasks. However, we relax this requirement to a 90\% approval rate for Stage III as it is simpler than the other annotation tasks.

\paragraph{Reward} The participants will be rewarded \$0.5 for each task in Stage I and II, and \$0.2 for the QA tasks in Stage III, with a possibility of a bonus depending on the overall quality. We actively monitor the response quality and send bonuses/rejections daily. Note that we collected 5 responses for each task in all three stages. 

\paragraph{Task lifetime} We set the lifetime as 10 days for tasks in Stage I and 20 days for those in Stage II and III. However, we found most of the tasks can be completed in less than 7 days.

\section{Dataset details}

\subsection{More statistics}\label{sec:more_stat}

\begin{figure}
    \centering
    \includegraphics[width=\linewidth]{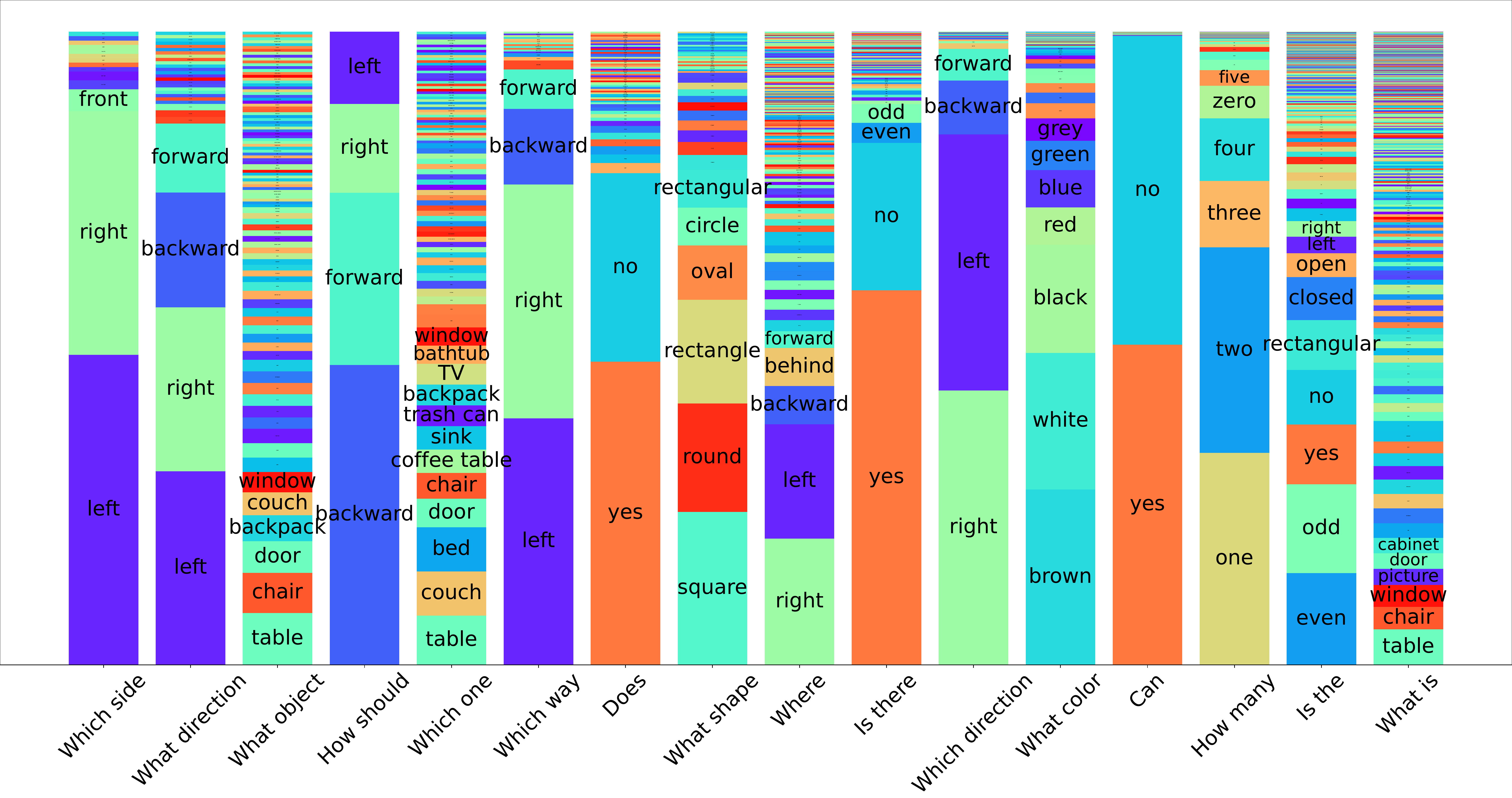}
    \caption{Answer distribution (organized by question prefixes) before balancing.}
    \label{fig:ans_dist_before}
\end{figure}

\begin{figure}
    \centering
    \includegraphics[width=\linewidth]{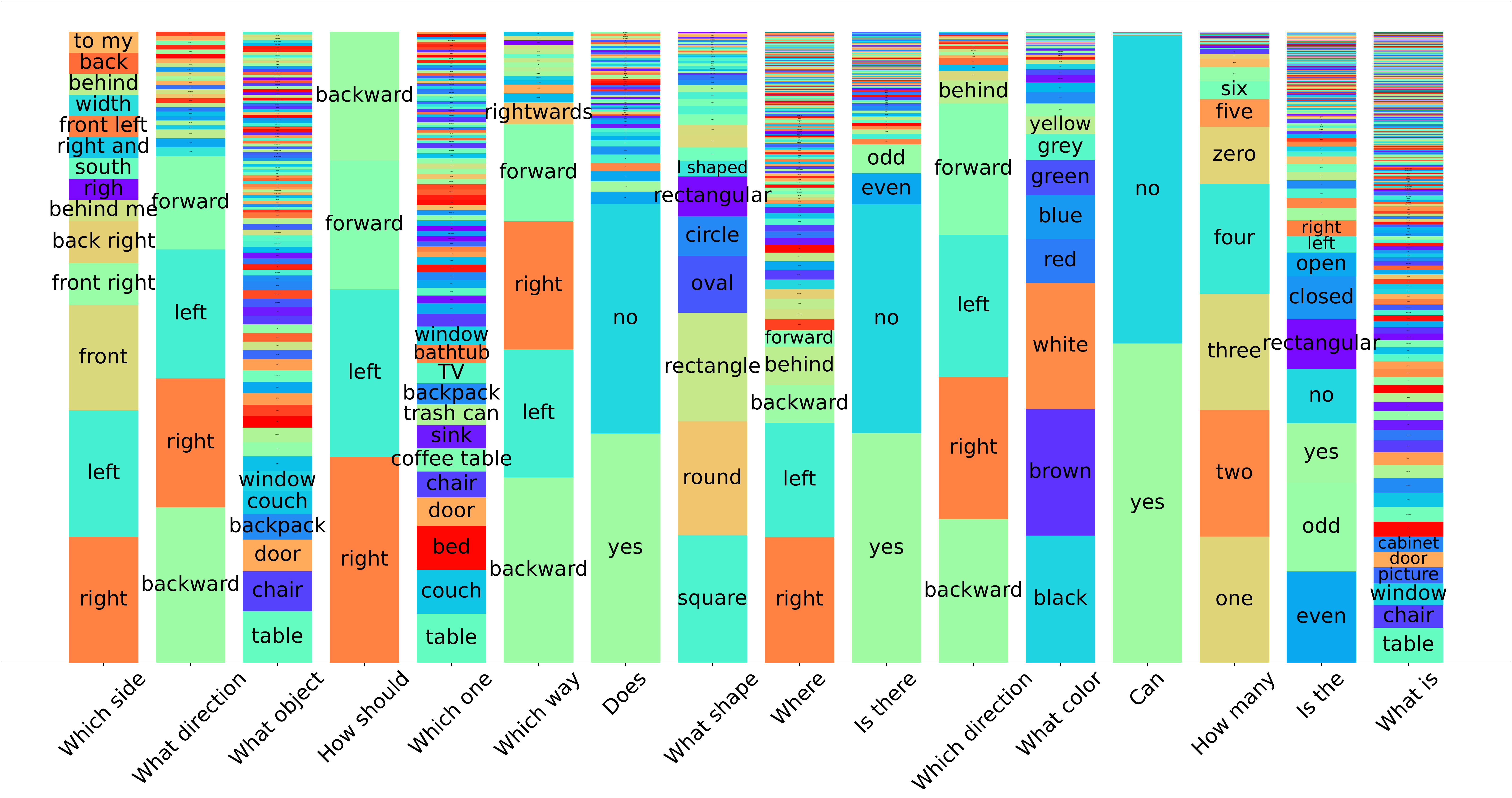}
    \caption{Answer distribution (organized by question prefixes) after balancing.}
    \label{fig:ans_dist_after}
\end{figure}

We provide the histogram of the answer distribution before \& after balancing in~\autoref{fig:ans_dist_before} and~\autoref{fig:ans_dist_after}, respectively. It can be seen that we manage to ensure there is no single answer that dominates any type of question (categorized by their prefixes). However, we do acknowledge that prefix-based balancing might still not be sufficient since models could also learn to use the n-grams pattern. A more effective avenue is collecting more questions with less-frequent answers, which we leave as future work. 

\begin{figure}
     \centering
     \begin{subfigure}[b]{0.45\textwidth}
         \centering
         \includegraphics[width=\textwidth]{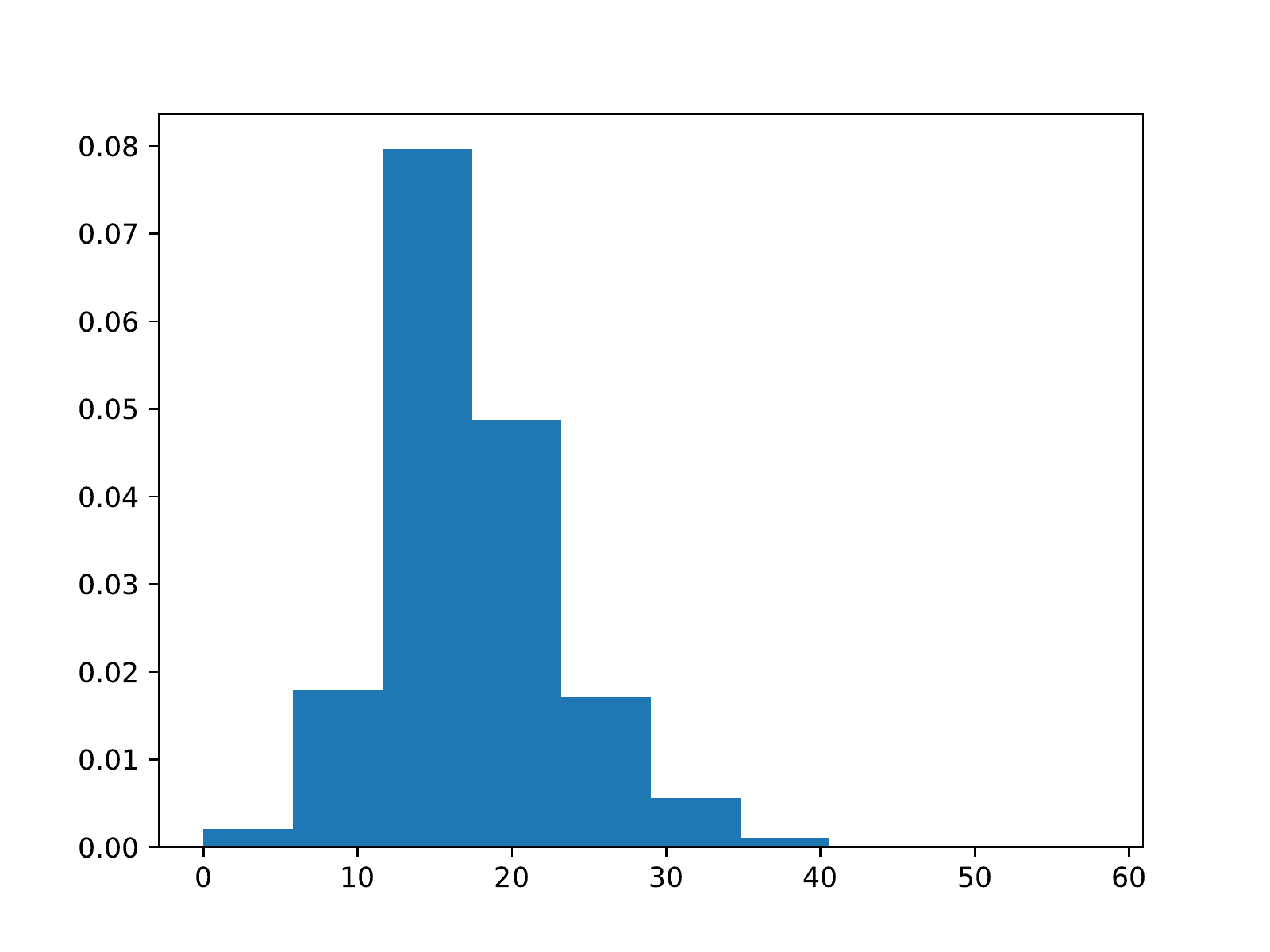}
         \caption{Histogram of situation description $\stxt$ length.}
         \label{fig:s_len_dist}
     \end{subfigure}
     \hfill
     \begin{subfigure}[b]{0.45\textwidth}
         \centering
         \includegraphics[width=\textwidth]{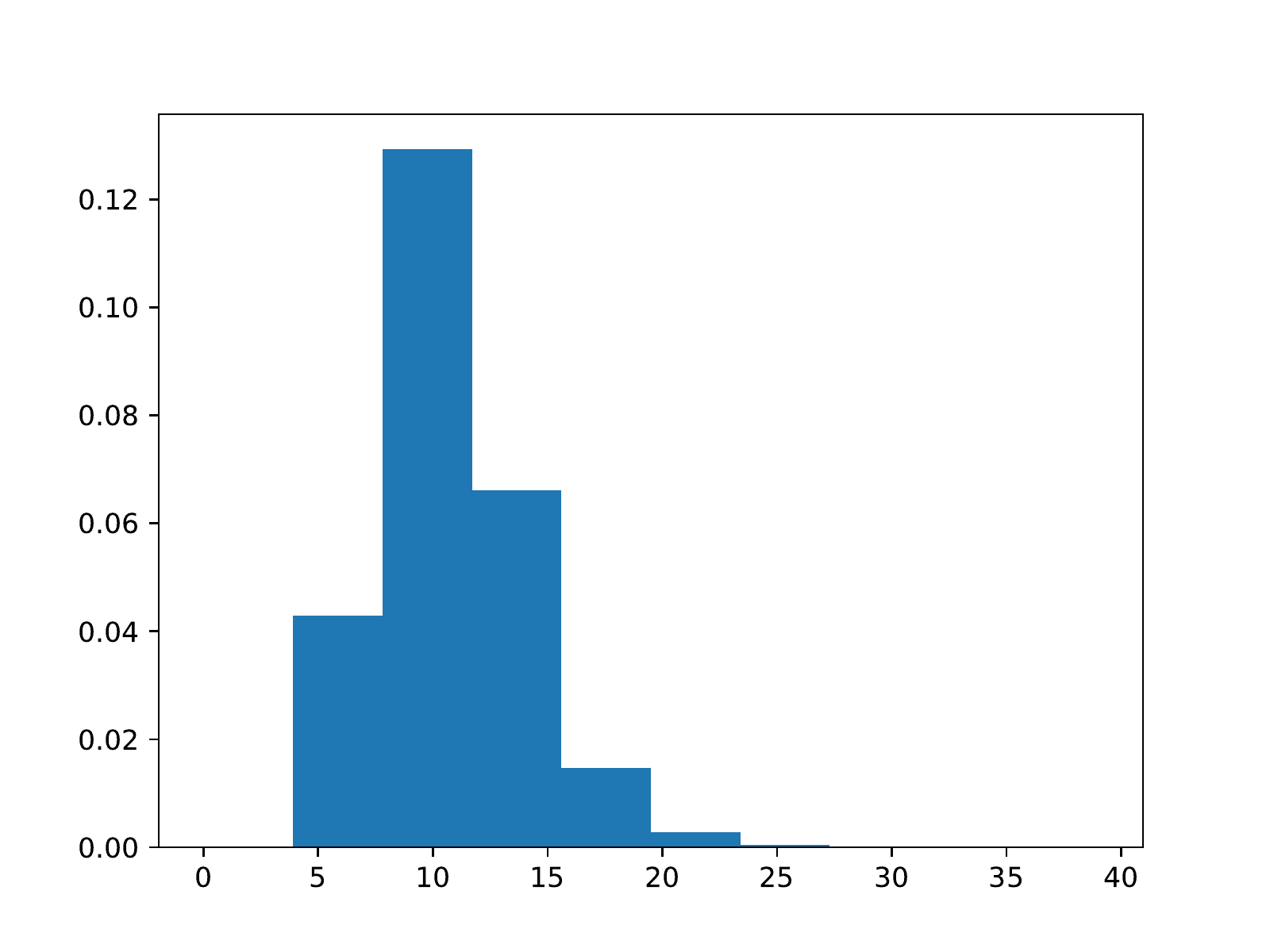}
         \caption{Histogram of question $q$ length.}
         \label{fig:q_len_dist}
     \end{subfigure}
\end{figure}

In~\autoref{fig:s_len_dist} and~\autoref{fig:q_len_dist}, we show the histogram of the length of situation description $\stxt$ and question $q$. Overall most of the descriptions and questions are middle-length sentences (10-20 words).

\subsection{Details on egocentric video and BEV image}
For egocentric videos, we uniformly downsample the frames of the original ScanNet \citep{dai2017scannet} video by using the first frame of every 20 frames. Afterward, we resize all the frames to $224\times 224$ to create the video used for training ClipBERT\citep{lei2021less}. Blender is used for rendering all BEV images. We compute the radius of the bounding sphere of the scene and put the camera at the top of the scene with a distance of 7 times the radius to the center of the bounding sphere. Images of size $1920\times 1080$ are rendered for clarity while the input to the MCAN\citep{yu2019deep} model is the resized version of the images to $224\times 224$.

\section{Model details}

\subsection{Input pipeline}
We follow the input pipeline in ScanQA\citep{azuma2022scanqa} without further modification. As for MCAN, we only transform the images to fit the ImageNet-pretrained encoder. In ClipBERT, we randomly sample 8 clips with each clip consisting of 2 frames of the video to feed into the model as the scene representation. Note that each frame is resized to $1000\times 1000$ following the practice of original ClipBERT\citep{lei2021less}.

\subsection{Hyper-parameters}

We provide the hyper-parameters of the considered models in~\autoref{tab:alg_param}.

\subsection{Additional details on zero-shot models}

We uniformly sample 30 sentences from our 3D caption sources for both models. When testing with the $\text{Unified QA}_{\text{Large}}$ model, we employ a simple greedy sampling method and the following prompt:
\begin{table}[H]
\begin{tabular}{|p{38em}|}
\hline
\makecell[tl]{\\\{$\stxt$\}\\\texttt{Q:} \{$q$\}\\\texttt{A:}\\} \\ \hline
\end{tabular}
\end{table}
, where $\{\stxt\}$ and $\{q\}$ are replaced by the situation description and question. For GPT-3, we use the \texttt{text-davinci-002} variant and the following prompt:
\begin{table}[H]
\begin{tabular}{|p{40em}|}
\hline
\makecell[tl]{\\\texttt{Context: There is a book on the desk. A laptop with a green cover} \\ \texttt{is to the left of the book.} \\ \texttt{Q: I'm working by the desk. What is on the desk beside the book?} \\ \texttt{A: laptop} \\ \texttt{Context:}\{$\stxt$\}\\\texttt{Q:} \{$q$\}\\\texttt{A:}\\} \\ \hline
\end{tabular}
\end{table}
, where we use a 1-shot example to demonstrate the format of our task. Interestingly, we found only GPT-3 would benefit from few-shot examples.
\subsection{Additional details on SCANQA/MCAN/CLIPBERT}

\textbf{ScanQA}~\citep{azuma2022scanqa}. We slightly modify the original ScanQA code base (from https://github.com/ATR-DBI/ScanQA) to make it fit our task better. The original reference branch is discarded and the supervision signal for the language classification branch is changed to make use of it as a regression branch. More specific details can be found below.
\begin{itemize}
            \item The original data loader only outputs the question as a whole (meaning that the situation is concatenated before the question), while our version split the two sentences.
            \item The original model takes language as 1 input, while we feed situation and question separately into the model.
            \item The original model uses 1 self-attention block and 1 cross-attention block for the fusion of language and visual features, while our version uses 2 self-attention blocks and 2 cross-attention blocks to treat situations and questions separately.
            \item The original model uses additive operation to fuse language \& visual features, while our version uses concatenation for fusion.
            \item To conduct the ablation experiment of blind test, we simply discard the output feature of VoteNet and only feed the situation feature and question feature into the QA head.
            \item To conduct the ablation experiment of w/o $\stxt$, we replace situation with several $\left\langle unk\right\rangle$ tokens to make a fair comparison.
            \item To add an auxiliary task into training, we change the supervision of the language classification head from Cross Entropy to MSE Loss to make it a regression head.
\end{itemize}

\textbf{MCAN}~\citep{yu2019deep}. We use the code base from RelVIT~\citep{ma2022relvit} (https://github.com/NVlabs/RelViT) since its implementation of MCAN could take raw images as input while the original one cannot. The default training setting is kept except for learning rate decay. We cancel it to make a fair comparison with the other baselines. We concatenate the situation before the question to make them as a whole and use this new sentence as the question that MCAN requires.

\textbf{ClipBERT}~\citep{lei2021less}. We use the official repository of ClipBERT (https://github.com/jayleicn/ClipBERT) and follow the instruction to transform our data into the format ClipBERT takes. The configuration file for MSR-VTT QA~\citep{xu2016msr} is used for generality as we find all the configuration files to be almost identical. The evaluated question types are changed since our focus is different from MSR-VTT. We turn off mixed precision training as we observe instability when using it. We concatenate the situation before the question to make them as a whole and use this new sentence as the question that ClipBERT requires.

\section{Additional empirical results}
\begin{figure}
    \centering
    \includegraphics[width=\linewidth]{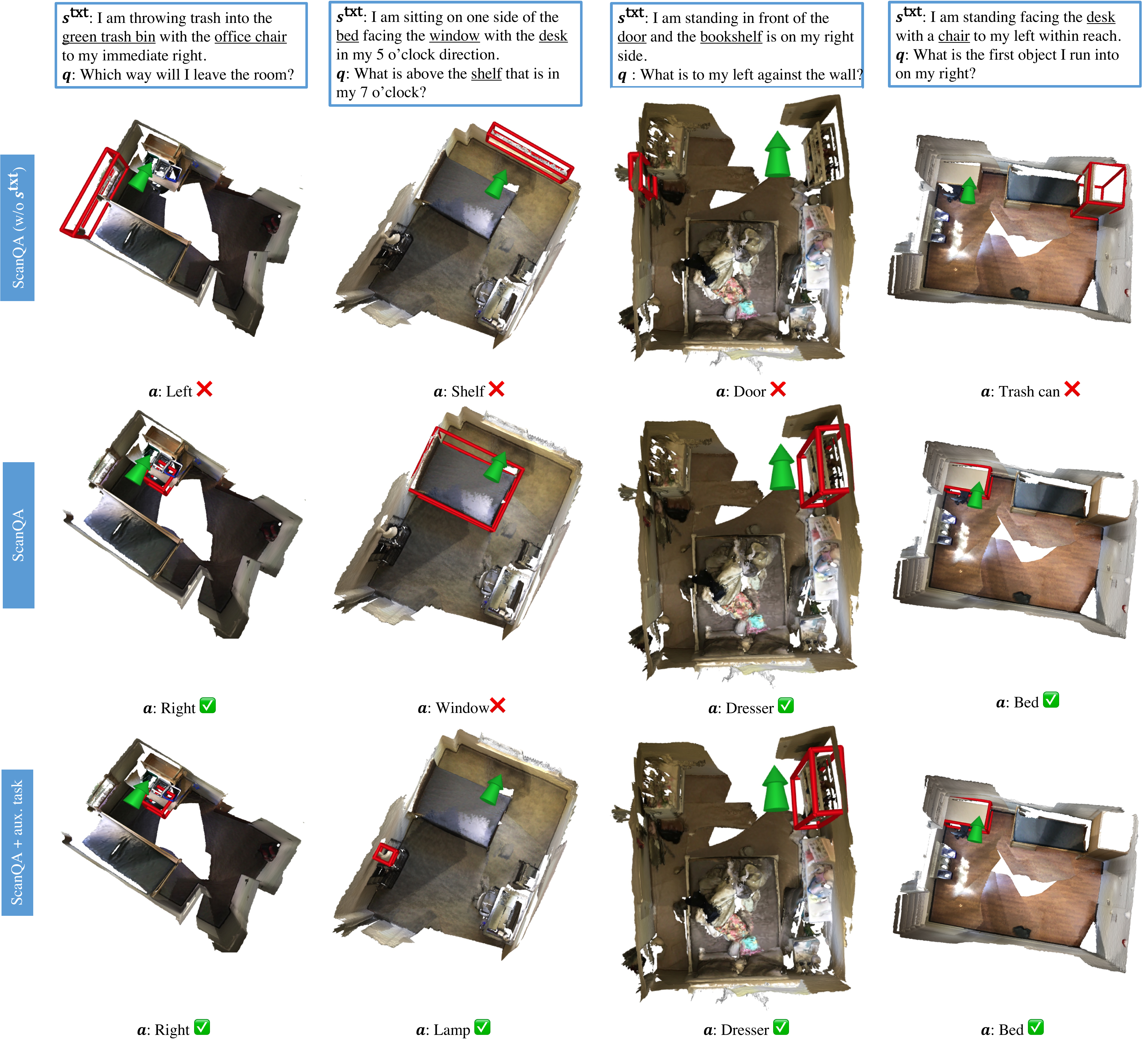}
    \caption{Additional qualitative results.}
    \label{fig:attn_more}
\end{figure}

\begin{figure}
    \centering
    \includegraphics[width=\linewidth]{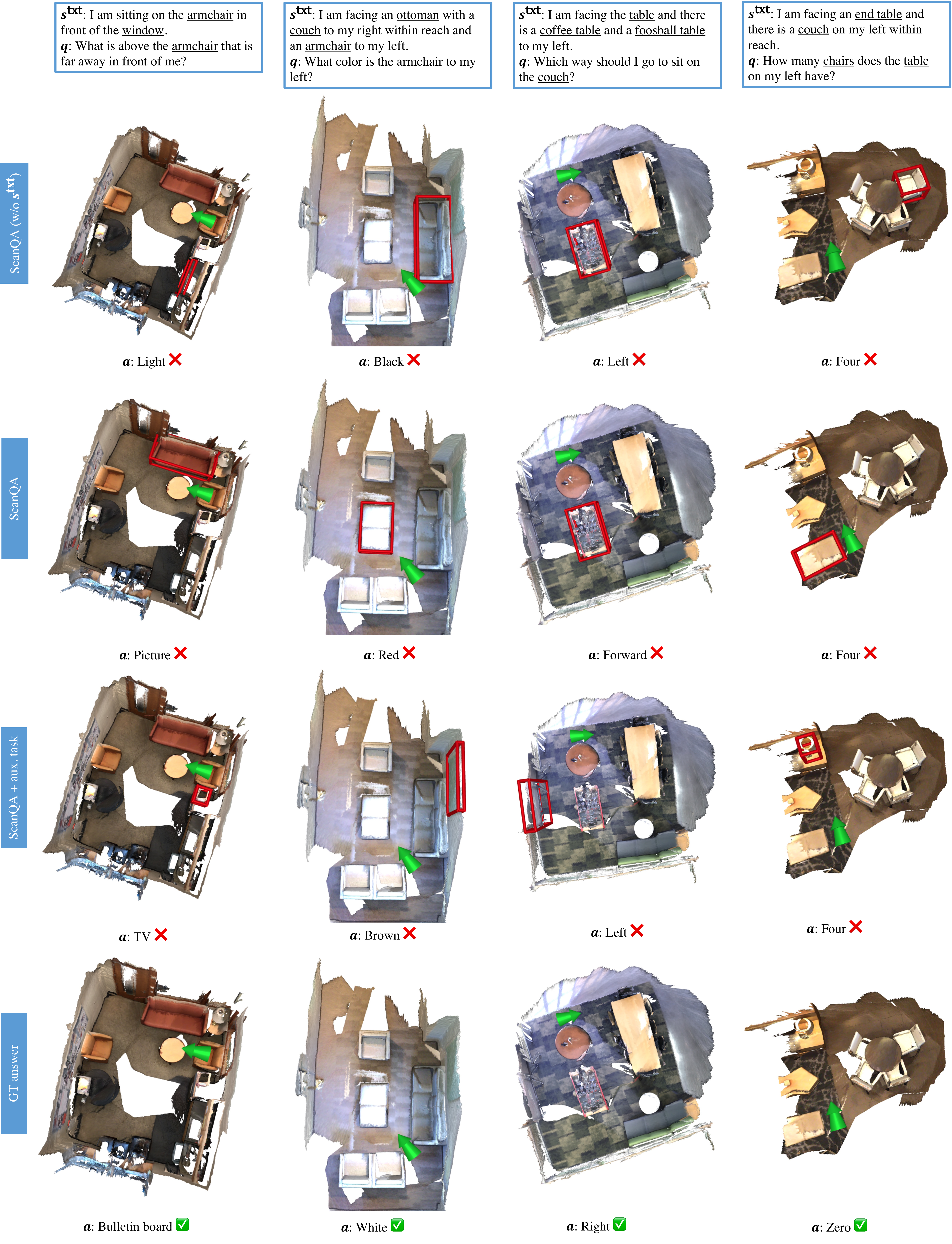}
    \caption{\textbf{Failure mode}. Models are likely to predict the wrong answers when they do not attend to relevant objects.}
    \label{fig:failure_mode}
\end{figure}

We provide additional qualitative results and failure modes in~\autoref{fig:attn_more} and~\autoref{fig:failure_mode}.

\section{Limitation and potential impact}
\paragraph{Limitations} One major limitation of \ac{sqa3d} is the selection of 3D scenes. Since our dataset is collected on mostly indoor ScanNet scenes about household environments, it cannot cover outdoor scenes and other types of scenes, ex. warehouse. This could limit the application to autonomous driving and warehouse robots, which are likely deployed to the scene types that do not present in \ac{sqa3d}. Moreover, all the scenes in ScanNet are static, \ie the agent cannot interact with the object, making the exploration in \ac{sqa3d} limited to hovering over the 3D scenes. However, many embodied tasks also require non-trivial interaction with articulated objects, \eg drawers. Therefore, the embodied scene understanding capability examined by \ac{sqa3d} can also be limited to non-interactive scenarios, \ie \textbf{situation understanding} and \textbf{situated reasoning}.

\paragraph{Societal impact} \ac{sqa3d} offers two sets of annotations: situations $\langle \stxt, \spos, \srot\rangle$ and QA $\langle q, a\rangle$. The situation annotations themselves could enable many exciting applications including building a real-world household assistant robot -- one of its core capabilities is connecting natural language instructions/descriptions to the situations in the scene, \eg locations. Moreover, non-trivial commonsense reasoning is also required in this process. Our annotations with accurate description-location pairs and the requirement of commonsense knowledge in text understanding could support these needs. The QA tasks also examine a wide spectrum of capabilities of embodied agents in household domains, making it a great benchmark for testing these household assistant robots. Finally, we will also release all the annotation interfaces and meta information, inviting everyone from either academia or industry to develop a customized version of QA datasets upon \ac{sqa3d} and its infrastructure, which might help with the development of the 3D-language-related research and products.

\begin{table}[h]
\renewcommand{\arraystretch}{1.1}
\centering
\caption{Hyper-parameters for the considered models.} 
\label{tab:shared_params}
\vspace{1mm}
\begin{tabular}{l l| l }
\toprule
\multicolumn{2}{l|}{Parameter} &  Value\\
\midrule
\multicolumn{2}{l|}{\it{ScanQA(modified)}}& \\
& Optimizer & Adam \\
& Gradient clipping norm & $1.0$ \\
& Epsilon &  $1e^{-8}$\\
& Weight decay factor & $1e^{-5}$ \\
& Beta hyperparameters for Adam & $[0.9, 0.999]$ \\
& Learning rate & $5e^{-4}$\\
& Learning rate schedule & No learning rate schedule \\
& Batch size & 16 \\
& Total training epochs & 50 \\
& Number of layers for transformer & 2 \\
& Number of heads for transformer & 8 \\
& MLP hidden size in MCAN& 256 \\
& MCAN flatten output size & 512 \\
& Model hidden size & 256 \\
& Number of VoteNet output proposals & 256 \\
& Position regression loss weight $\alpha$ for auxiliary task& 1.0 \\
& Rotation regression loss weight $\beta$ for auxiliary task& 1.0 \\

\midrule
\multicolumn{2}{l|}{\it{MCAN}}& \\
& Optimizer & AdamW \\
& Gradient clipping norm & $0.5$ \\
& Epsilon &  $1e^{-8}$\\
& Weight decay factor & $0$ \\
& Beta hyperparameters for Adam & $[0.9, 0.999]$ \\
& Learning rate & $1e^{-4}$\\
& Learning rate schedule & No learning rate schedule \\
& Batch size & 16 \\
& Total training epochs & 12 \\
& Number of layers for transformer & 6 \\
& Number of heads for transformer & 8 \\
& MLP hidden size in MCAN& 512 \\
& MCAN flatten output size & 1024 \\
& Model hidden size & 512 \\

\midrule
\multicolumn{2}{l|}{\it{ClipBERT}}& \\
& Optimizer & AdamW \\
& Gradient clipping norm & $5.0$ \\
& Epsilon &  $1e^{-6}$\\
& Weight decay factor & $1e^{-3}$ \\
& Beta hyperparameters for Adam & $[0.9, 0.98]$ \\
& Learning rate & $5e^{-5}$\\
& Learning rate schedule & No learning rate schedule \\
& Batch size & 16 \\
& Total training epochs & 10 \\

\bottomrule
\label{tab:alg_param}
\end{tabular}
\end{table}
\end{document}